%% file: main.tex
\documentclass{article}

\usepackage[main, final]{neurips_2026}

\usepackage[utf8]{inputenc}
\usepackage[T1]{fontenc}

\usepackage[table]{xcolor}
\usepackage{url}
\usepackage{booktabs}
\usepackage{amsfonts}
\usepackage{nicefrac}
\usepackage{microtype}
\usepackage{graphicx}
\usepackage{multirow}
\usepackage{wrapfig}
\usepackage{adjustbox}
\usepackage{diagbox}
\usepackage{placeins}
\usepackage[accsupp]{axessibility}
\usepackage[most]{tcolorbox}
\usepackage{subcaption} 
\usepackage{hyperref}
\usepackage{cleveref}

\newcommand{\rtext}[1]{\textcolor{red}{(#1)}}

\definecolor{revcolor}{RGB}{0, 112, 192}

\title{When Prompts Override Vision: Prompt-Induced Hallucinations in LVLMs}

\author{
Pegah Khayatan$^1$ \quad
Jayneel Parekh$^1$ \quad \quad  \quad Arnaud Dapogny$^1$ \\ \textbf{Mustafa Shukor}$^1$ \quad \quad \textbf{Alasdair Newson}$^1$ \quad \quad \textbf{Matthieu Cord}$^{1,2}$\\ 
$^1$ISIR, Sorbonne Université, Paris, France \quad $^2$Valeo.ai, Paris, France\\
\texttt{\{pegah.khayatan\}@sorbonne-universite.fr}
}
\begin{document}

\setlength{\textfloatsep}{8pt}
\setlength{\floatsep}{6pt}
\setlength{\intextsep}{6pt}
\setlength{\abovecaptionskip}{4pt}
\setlength{\belowcaptionskip}{0pt}

\maketitle

\input{icml2026_files/1_abstract}

\input{icml2026_files/1_introduction}

\input{icml2026_files/1.5_related_works}

\input{icml2026_files/2_text_induced_hallucination}

\input{icml2026_files/3_solution}

\input{icml2026_files/3_preliminary}

\input{icml2026_files/3_data_synthesis}

\input{icml2026_files/6_experiments}

\input{icml2026_files/7_ablation}
\input{icml2026_files/conclusion}

\begin{ack}
This work has been partially supported by ANR grant VISA DEEP (ANR-20-CHIA-0022), HPC resources of IDRIS under the file A0191016602 allocated by GENCI, and Cluster PostGenAI@Paris (ANR-23-IACL-0007, FRANCE 2030).
\end{ack}

\bibliographystyle{unsrt}   %
\bibliography{bibliography}

\appendix

\section*{Supplementary Material}

\input{icml2026_files/appendix}   %

\end{document}

%% file: icml2026_files/1_abstract.tex
\begin{abstract}
    \textit{Despite impressive progress in capabilities of large vision-language models (LVLMs), these systems remain vulnerable to hallucinations, i.e., outputs that are not grounded in the visual input. Prior work has attributed hallucinations in LVLMs to factors such as limitations of the vision backbone or the dominance of the language component, yet the relative importance of these factors remains unclear. To resolve this ambiguity, we propose \textbf{\textit{HalluScope}}, a benchmark to better understand the extent to which different factors induce hallucinations.
    Our analysis indicates that hallucinations largely stem from excessive reliance on textual priors and background knowledge, especially information introduced through textual instructions.
    To mitigate hallucinations induced by textual instruction priors, we propose \textbf{HalluVL-DPO}, a framework for fine-tuning off-the-shelf LVLMs towards more visually grounded responses. HalluVL-DPO leverages preference optimization using a curated training dataset that we construct, guiding the model to prefer grounded responses over hallucinated ones.
    We demonstrate that our optimized model effectively mitigates the targeted hallucination failure mode, while preserving or improving performance on other hallucination benchmarks and visual capability evaluations.
    Our code and data are publicly available. \footnote[1]{Github page: \url{https://github.com/pegah-kh/HalluScope}}$^,$ \footnote[2]{Blog/Project page: \url{https://pegah-kh.github.io/projects/prompts-override-vision/}} 
    }

\end{abstract}

%% file: icml2026_files/1_introduction.tex
\section{Introduction}

Large Vision–Language Models (LVLMs) have recently achieved remarkable progress across multimodal tasks such as image captioning, visual reasoning, and instruction following \cite{liu2023llava, wang2024qwen2, laurenccon2024matters, wang2025internvl3, bai2025qwen3}. Despite these advances, they are still prone to visual hallucinations \cite{Sun2023AligningLM,shukor2024beyond,bai2024hallucination,huang2024opera}, i.e., generating outputs that are not grounded in the input visual information. Such failures limit the deployment of LVLMs in safety-critical applications such as autonomous systems and medical diagnosis.
While even standalone 
visual encoders can exhibit object hallucinations \cite{liu-etal-2024-investigating,li-etal-2025-visual},
emerging evidence suggests a different failure mode: even when visual recognition is accurate, modern LVLMs frequently generate incorrect responses, reflecting the dominance of the text modality over the visual modality \cite{xie2024v, leng2024mitigating, Baldassini_2024_CVPR}.
In such cases the output is shaped more by learned co-occurrence priors and the text instruction than by the image itself. This points to a shift in failure modes: as visual backbones improve, hallucinations increasingly arise from conflicts between language priors and visual information rather than from perceptual limitations alone.
However, existing evaluation benchmarks including POPE \cite{DBLP:conf/emnlp/LiDZWZW23}, CHAIR \cite{Rohrbach2018ObjectHI}, SHR \cite{zhao2023hallucinations}, and MMHAL-Bench \cite{Sun2023AligningLM} do not distinguish between hallucinations originating %
from perception failures, learned object co-occurrence priors, or presuppositions introduced by the instruction itself. As a result, the relative contribution of these factors remains poorly understood, limiting both diagnosis and mitigation.

\begin{figure}[ht]
\centering
\includegraphics[width=0.98\linewidth]{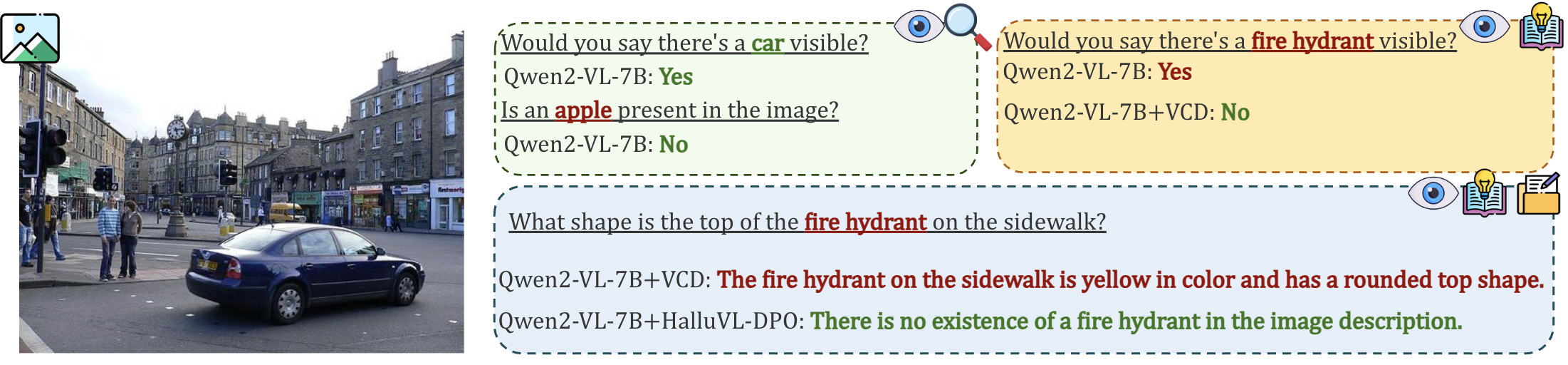}
\caption{\textbf{Vision–language hallucination failure modes and mitigation.} The model reliably recognizes present objects or the absence of random objects, but tends to hallucinate plausible yet absent (i.e., adversarial) objects, especially when the instruction presupposes their presence. Our \textit{HalluVL-DPO} framework substantially mitigates this failure mode.}
\label{fig:failure_modes}
\end{figure}

To address this limitation, we introduce \textit{HalluScope}, a diagnostic benchmark to
isolate distinct 
causes of hallucination in LVLMs. Rather than treating hallucination as a single phenomenon, \textit{HalluScope} separates errors arising from (i) visual perception limitations, (ii) reliance on learned semantic priors, and (iii) reliance on presuppositions introduced by the instruction. Using this benchmark, we demonstrate that hallucinations in modern LVLMs are predominantly driven by textual instructions, even when visual perception remains reliable (\Cref{fig:failure_modes}). These findings suggest that improving LVLM reliability is increasingly an alignment problem between modalities.

To mitigate hallucinations induced by text instruction presuppositions, we propose \textit{HalluVL-DPO}, a framework for fine-tuning off-the-shelf LVLMs. \textit{HalluVL-DPO} employs a sample-{\linebreak}informativeness based variant of preference-optimization \cite{rafailov2023direct}, and leverages a large-scale preference dataset we construct specifically to favor grounded over hallucinated responses. Models fine-tuned by our framework effectively reduce the targeted hallucinations, achieve consistent improvements on hallucination benchmarks, and maintain strong general multimodal performance.
Our contributions are summarized as follows: 
\begin{itemize}
\item \textit{A benchmark for assessing the impact of distinct causes of LVLM hallucinations.} We introduce \textit{HalluScope}, a benchmark designed to disentangle distinct causes of hallucination: perception failures, learned object co-occurrence priors, or presuppositions introduced by the instruction. Using \textit{HalluScope}, we show that hallucinations in modern LVLMs predominantly arise from over-reliance on textual instruction presuppositions and learned semantic priors rather than limitations of visual perception, revealing a shift in failure modes as visual backbones improve.
\item \textit{A framework to effectively mitigate hallucinations caused by over-reliance on textual instruction presuppositions.} We propose \textit{HalluVL-DPO}, an LVLM fine-tuning framework based on sample-informativeness based preference optimization \cite{rafailov2023direct}. Our framework explicitly encourages visually grounded responses over hallucinated answers, using a challenging, generated training dataset. Our approach effectively reduces targeted hallucinations and shows consistent gains on hallucination benchmarks.

\end{itemize}
We publicly release our HalluScope benchmark, its construction pipeline, and all components of HalluVL-DPO, including the training data, dataset construction pipeline, and fine-tuned model checkpoints.

%% file: icml2026_files/1.5_related_works.tex
\section{Related work}
\noindent \textbf{Evaluating Hallucinations in LVLMs.}
Hallucinations in LVLMs manifest in several forms. In open-ended text generation, hallucination commonly refers to producing content that is not grounded in the visual input. This is typically evaluated using metrics such as CHAIR \cite{Rohrbach2018ObjectHI}, and more recently SHR \cite{zhao2023hallucinations} and ALOHa \cite{petryk2024aloha}. Another form of hallucination arises when the model generates incorrect attributes for objects in the image, as evaluated by benchmarks such as MMHalBench \cite{Sun2023AligningLM}.
In closed-ended settings, where prompts query the presence or properties of specific objects, hallucinations are defined as incorrect responses. This form is commonly assessed using benchmarks such as POPE \cite{DBLP:conf/emnlp/LiDZWZW23} and HallusionBench \cite{guan2024hallusionbench}. 
While these benchmarks measure outcome-level correctness, they provide limited insight into the relative importance of different causes of hallucinations. \newline
\noindent \textbf{Mitigating Hallucinations in LVLMs.}
Previous studies to address hallucinations can be grouped into two main categories: \textit{post-hoc (training-free)} methods and \textit{optimization-based} methods. 
Training-free approaches intervene at different points: manipulating attention, by targeting
``hallucination attention heads'' \cite{yang2025mitigatinggg}, re-distributing it away from
semantically uninformative \textit{sink} tokens \cite{kang2025see}, or amplifying visual tokens
\cite{liu2407paying, an2024agla, cho2025do}; steering or probing internal representations
\cite{liu2024reducing, parekh2025learning, yang2025hallucination, wangdamo, wang2024mllm}; or
contrasting output distributions between the original and a noised image \cite{leng2024mitigating} or between visual and text-only inputs \cite{liu2407paying}. Such methods typically increase inference time \cite{leng2024mitigating} or require modifications to the model's implementation \cite{yang2025mitigatinggg}, and are inherently limited to the hypotheses they consider \cite{kang2025see}.
Optimization-based methods address hallucinations more fundamentally by directly adjusting the LVLM’s weights, allowing the model to systematically align its outputs with grounded visual information and learned preferences.
\cite{liu2024mitigating} proposes a large-scale instruction-tuning dataset including samples with misleading and hallucination-inducing instructions. %
Datasets used in these works rely on closed-source vision models to detect hallucinations \cite{yang2025mitigating}, or combine them with richly annotated datasets like Visual Genome \cite{zhao2023hallucinations}, limiting reproducibility and scalability. \cite{peng2025mitigating} does not release training data, further hampering reproducibility. Crucially, these methods target object hallucination broadly rather than the textually-induced presupposition failures we focus on. \\
Our pipeline addresses this failure more directly through two methodological contributions that distinguish our approach from prior DPO-based work: sample-informativeness-based weighting of preference pairs, and data construction strategies designed to produce challenging, grounded preference signals. We include two recent DPO-based approaches, SENTINEL \cite{peng2025mitigating} and OPA-DPO \cite{yang2025mitigating}, as well as a reimplementation of Antidote \cite{wu2025antidote}, as direct baselines.

%% file: icml2026_files/2_text_induced_hallucination.tex
\section{\texorpdfstring{\textit{HalluScope}\includegraphics[height=1.2em]{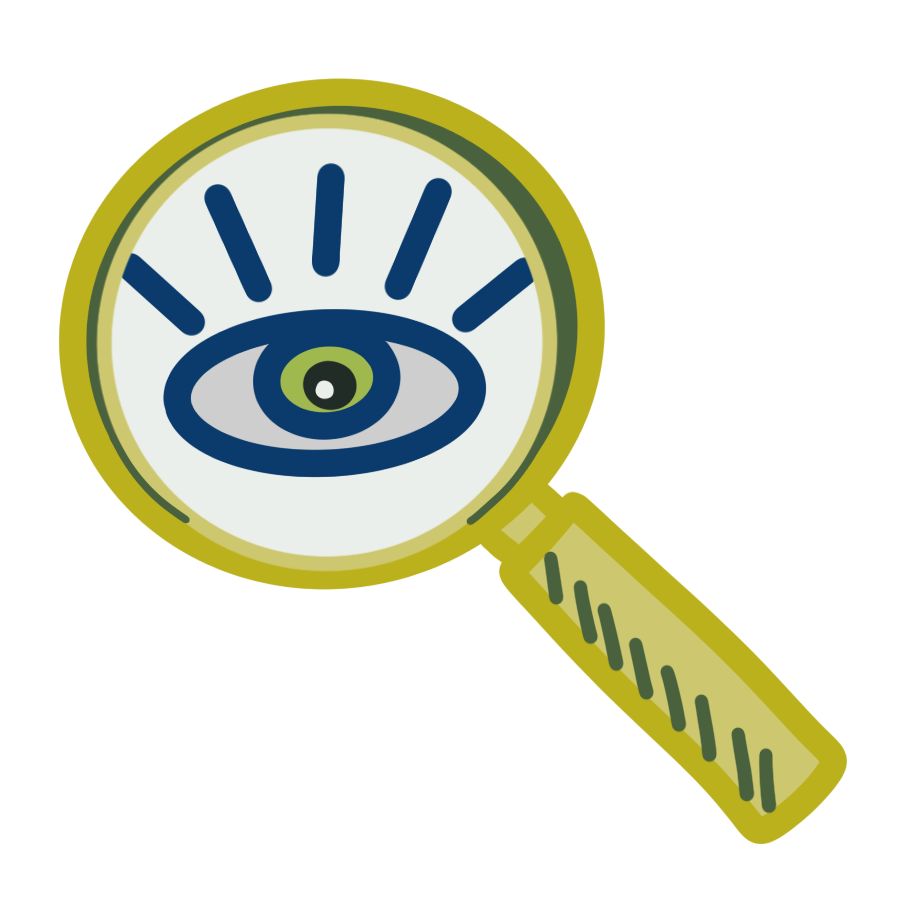}}{HalluScope}: Diagnosing Hallucinations in LVLMs}
\label{main:halluscope}

While existing benchmarks \cite{Rohrbach2018ObjectHI, zhao2023hallucinations, petryk2024aloha, Sun2023AligningLM, DBLP:conf/emnlp/LiDZWZW23, guan2024hallusionbench} assess hallucination, they do not disentangle different contributing factors, such as visual perception errors, the model’s prior knowledge about object co-occurrence, or textual presuppositions.  
\cite{seth2024towards} is a step in this direction, but varies question type and difficulty simultaneously, making causal attribution difficult.
To address this gap, 
we propose \textit{HalluScope}, a benchmark designed to systematically assess how different causes influence hallucinations.
The benchmark comprises a diverse set of images, each paired with questions to evaluate distinct hallucination factors:\\
-- We evaluate the visual perception of the model through \textit{Positive Object Recognition questions}, querying the presence of objects visually present in the image, and \textit{Random Object Recognition questions}, querying the presence of absent objects that are unrelated to the scene.\\
-- Next, we measure the extent of hallucinations induced when the model relies on learned object co-occurrence, instead of visual evidence, through \textit{Adversarial Object Recognition questions}. These questions query the presence of semantically plausible but visually absent objects. For example, if an image shows a table but no chair, the model may still answer “yes” when asked whether a chair is present, revealing reliance on learned co-occurrence biases rather than the image content.\\
-- Lastly, the \textit{Adversarial Presupposition Attribute} questions introduce an additional level of difficulty by querying an attribute (e.g., color) of an adversarial object whose presence is implicitly assumed in the question. This subset evaluates whether the model relies on the presuppositions in the textual instruction or on grounded visual evidence. For example, a question such as “What is the color of the chair?” implicitly assumes the chair’s presence and can induce the model to hallucinate one.

\subsection{Benchmark Construction}
\label{benchmark_construction}

\begin{figure}[h!]
    \centering

    \includegraphics[width=1\linewidth]{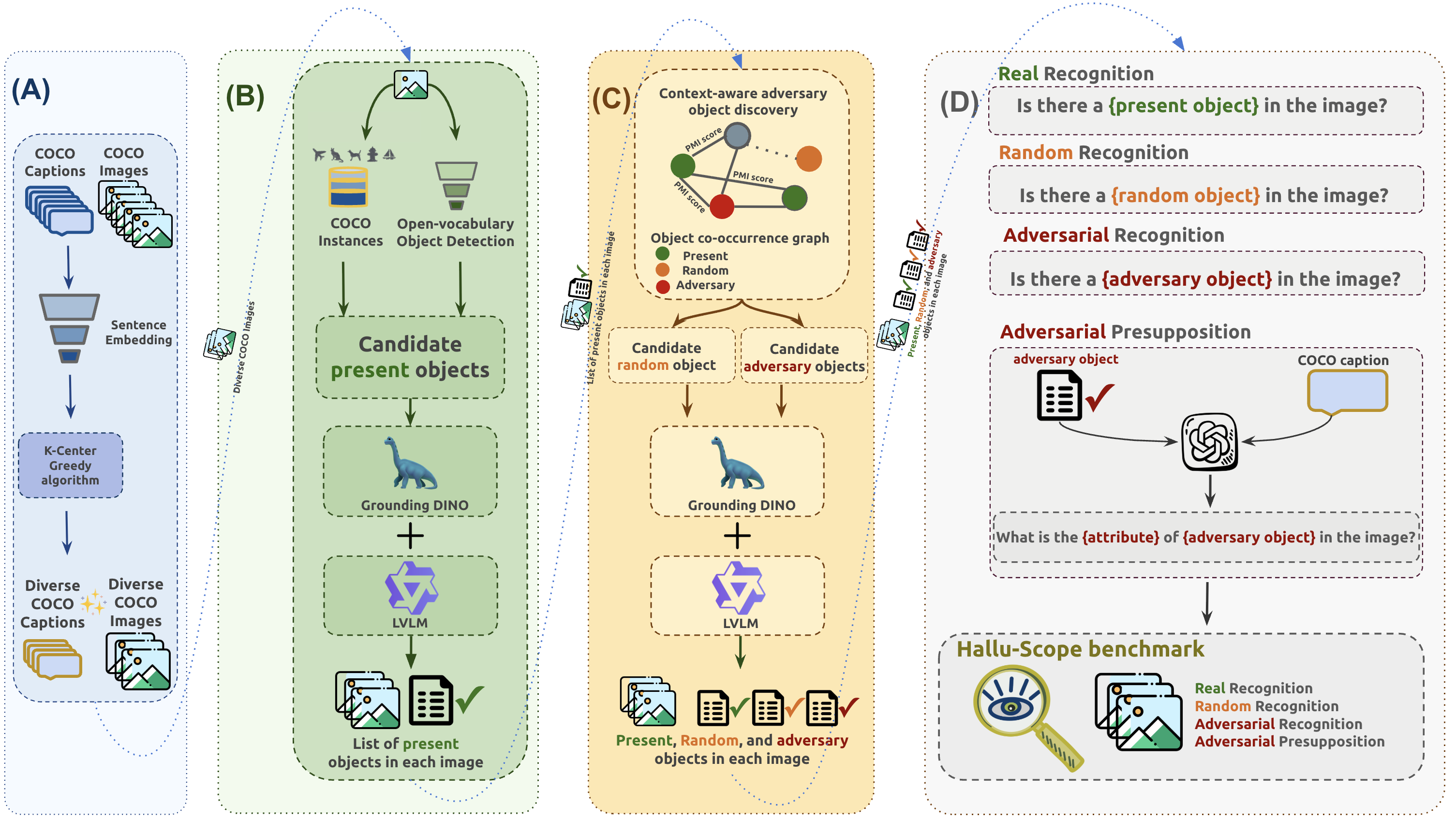}
    \vfill
    \includegraphics[width=1\linewidth]{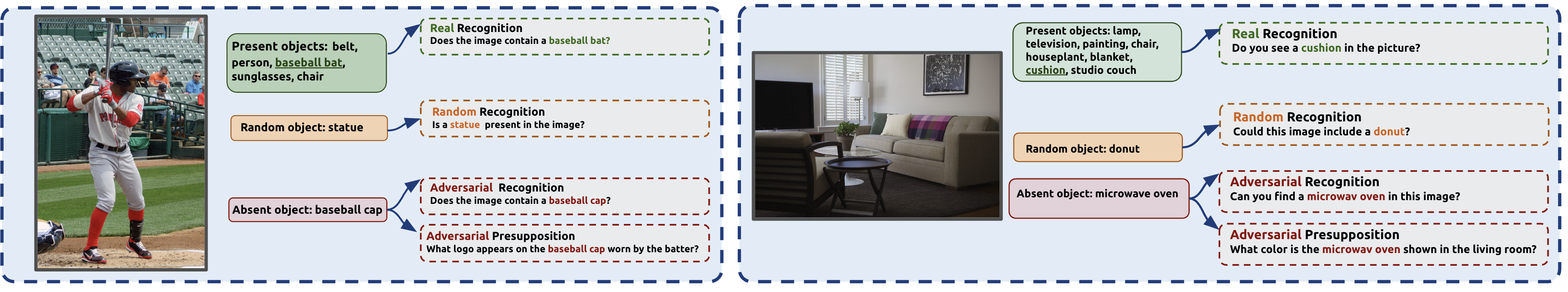}
    \caption{\textbf{\textit{HalluScope} construction pipeline.} (A) We start by constructing a semantically diverse subset of COCO images. (B) Objects present in each image are then detected and grounded. (C) Contextually plausible but visually absent adversarial objects are identified using object co-occurrence statistics, and each image is annotated with a present object, a random absent object, and an adversarial object. (D) Finally, different types of questions are generated for each image to probe models’ reliance on visual evidence versus textual priors.}
    \label{fig:aohd_construction}
\end{figure}

\textbf{A. Diverse Sample Pool.} We construct a semantically diverse subset of COCO images that covers a wide range of semantic content. To this end, we embed all captions using a sentence transformer and apply a $K$-Center Greedy algorithm to the resulting embeddings, iteratively selecting samples that are maximally distant from one another in embedding space. Because captions reflect the semantic content of images, this procedure yields a semantically diverse set of images. \newline
\textbf{B. Object Detection.}
We employ a robust pipeline to identify all objects present in each image within our selected sample pool. Candidate objects are sourced from \textit{(1) COCO instances annotations}, which provide high-quality, human-verified object labels, and \textit{(2) detections produced by an object detection model (Florence-2-large \cite{xiao2024florence})}, which expands coverage to a more diverse and fine-grained set of objects. To check for the presence or absence of a candidate object in an image, we use a \textit{two-stage verification pipeline}. \textit{(1) 
An object grounding model (Grounding-DINO \cite{liu2024grounding})} is first prompted with the name of each candidate object to obtain a grounding confidence score. When no score is returned, indicating that no corresponding region is detected, the object is labeled absent. Else, if the score exceeds a predefined threshold: it indicates object presence. For the remaining lower-confidence cases, 
\textit{(2) an LVLM} \cite{wang2024qwen2} is subsequently queried about the object’s presence. In the left example of \Cref{fig:aohd_construction}, the object detector initially identifies a baseball cap, sunglasses, chair, belt, baseball bat, and person; the verification stage subsequently removes the baseball cap from the set of present objects. \newline
\textbf{C. Object Co-occurrence and Context-aware Adversarial Object Discovery.}
Our goal is to annotate each image with three objects: a present $\mathcal{O}_{\text{present}}$, a random $\mathcal{O}_{\text{random}}$, and an adversarial $\mathcal{O}_{\text{adversarial}}$.
LVLMs tend to hallucinate objects that are contextually plausible but visually absent, due to learned co-occurrence patterns in training data \cite{lu2025mitigating, pi2024strengthening}.
We exploit this by building an object co-occurrence graph and ranking absent objects by their cumulative \textit{pairwise pointwise mutual information (PMI)} \cite{church-hanks-1990-word} score with objects present in the scene, selecting the highest-ranked absent candidate as $\mathcal{O}_{\text{adversarial}}$, the object most associated with it as $\mathcal{O}_{\text{present}}$, and a random absent object as $\mathcal{O}_{\text{random}}$.
All candidates are verified using the two-stage pipeline. Full details are in \Cref{app:object_cooccurrence}.
For example, in \Cref{fig:aohd_construction}, a baseball game scene with verified present objects (belt, sunglasses, baseball bat, person, chair) yields \textit{baseball cap} as the top absent adversarial candidate, \textit{baseball bat} as the paired present object (highest PMI with the adversarial object), and \textit{statue} as the random object.
\newline
\textbf{D. Question Generation.}
For each sample, we generate four questions: three \textit{object recognition} questions querying the presence of $\mathcal{O}_{\text{present}}$, $\mathcal{O}_{\text{random}}$, and $\mathcal{O}_{\text{adversarial}}$, and one \textit{adversarial presupposition attribute} question that concerns the adversarial object and implicitly assumes its presence. Recognition questions are instantiated by randomly sampling from a set of templates (e.g., ``Is a \{object\} present in the image?''), whereas the adversarial presupposition attribute question is generated via prompting \texttt{GPT5-mini}, conditioned on the image caption, $\mathcal{O}_{\text{present}}$ and $\mathcal{O}_{\text{adversarial}}$. \newline
For example, in the baseball scene shown in \Cref{fig:aohd_construction}, where $\mathcal{O}_{\text{present}}$ is \textit{baseball bat}, $\mathcal{O}_{\text{random}}$ is \textit{statue}, and $\mathcal{O}_{\text{adversary}}$ is \textit{baseball cap}, the resulting questions are: ``Does the image contain a baseball bat?'', ``Is a statue present in the image?'', ``Does the image contain a baseball cap?'', and ``What logo appears on the baseball cap worn by the batter?'' \newline 
HalluScope includes 3K images in each subset, where subsets are defined by the source of candidate objects (COCO Instances or Florence object detector).
Our \textit{object recognition} subset can be viewed as an extension of POPE \cite{DBLP:conf/emnlp/LiDZWZW23} but we include over an order of magnitude more images and by not restricting queried objects to a closed set of object categories. \newline
It should be noted that our pipeline is highly flexible: images can be drawn from any dataset, candidate objects can come from various sources, and verification can be performed using an ensemble of LVLMs to improve annotation quality even further.
\subsection{Results and findings}
\label{sec:halluscope_results}

\begin{table}[t]
\centering
\scriptsize
\setlength{\tabcolsep}{2pt}
\begin{tabular}{lcccc|cccc}
\toprule
\textbf{Model}
& \multicolumn{4}{c}{\textbf{Instances ($\%$) ($\uparrow$)}}
& \multicolumn{4}{c}{\textbf{Florence ($\%$) ($\uparrow$)}} \\
\cmidrule(lr){2-5} \cmidrule(lr){6-9}
& \textbf{Rec}$_{\text{pos}}$ & \textbf{Rec}$_{\text{rnd}}$ & \textbf{Rec}$_{\text{adv}}$ & \textbf{AdP}
& \textbf{Rec}$_{\text{pos}}$ & \textbf{Rec}$_{\text{rnd}}$ & \textbf{Rec}$_{\text{adv}}$ & \textbf{AdP} \\
\midrule
\addlinespace[2pt]
LLaVA-1.5-7B  & 94.8 & 91.1 & 54.6 \rtext{-38.4}  & 6.3 \rtext{-86.7} & 94.9 & 87.2  & 62.4 \rtext{-28.6} & 5.4 \rtext{-85.7} \\
Qwen2-VL-7B    & 92.5 & 97.9 & 86.8 \rtext{-8.4} & 59.3 \rtext{-35.9} & 94.1 & 95.2 & 83.8 \rtext{-10.9} & 56.7 \rtext{-38.1} \\
InternVL3-5-8B & 98.6 & 94.1 & 64.9 \rtext{-31.5} & 41.6 \rtext{-54.7} & 96.9 & 91.7 & 73.3 \rtext{-21.1} & 47.0 \rtext{-47.4} \\
Molmo-7B-D     & 97.5 & 85.6 & 61.2 \rtext{-30.3} & 29.1 \rtext{-62.5} & 97.7 & 76.5 & 59.5 \rtext{-27.5} & 23.9 \rtext{-63.2} \\
\bottomrule
\end{tabular}
\vspace{3pt}
\caption{
\textbf{Object recognition accuracy and Adversarial Presupposition (AdP) performance on HalluScope.} All models achieve high accuracy on positive and random recognition, but accuracy degrades on adversarial object recognition due to reliance on learned co-occurrence statistics. \noindent \textit{\textsc{A}dversarial Presupposition (AdP)} performance further drops, highlighting the role of textual instruction priors in driving hallucinations.
}
\label{tab:recognition_ap}
\end{table}
Performance on the recognition questions is measured by checking whether the response correctly indicates the presence or absence of the object (i.e., “yes” or “no”), and is reported as \textit{Positive Recognition} ($\text{Rec}_{\text{pos}}$), \textit{Random Recognition} ($\text{Rec}_{\text{rnd}}$), and  \textit{Adversarial Recognition} ($\text{Rec}_{\text{adv}}$). For the adversarial presupposition attribute questions, we report the \textit{\textsc{A}dversarial} 
\textit{\textsc{P}resupposition (AdP)}  performance, defined as the proportion of samples for which the model correctly rejects the false presupposition, introduced by the instruction, that the adversarial object is present in the image. This is assessed using an LLM that determines whether the model’s response denies or maintains the false assumption. Accuracy is reported for both subsets (more details in \Cref{app:evaluation_aohd}). The results of evaluating several models are shown in \Cref{tab:recognition_ap}. \newline
To estimate how often the visual backbone fails without introducing any specific information in the prompt about the adversarial object, we observe the $\text{Rec}_{\text{pos}}$ and $\text{Rec}_{\text{rnd}}$ performance. Across all models and both object sets, both accuracies remain consistently high, above 85$\%$. This indicates that visual backbones are generally good enough to identify the present and any random absent objects. \newline
To estimate how often models rely on learned co-occurrence statistics rather than visual evidence, we assess $\text{Rec}_{\text{adv}}$. Across all models, $\text{Rec}_{\text{adv}}$ is lower 
by 8.4$\%$ to 38.4$\%$ compared to the average of $\text{Rec}_{\text{pos}}$ and $\text{Rec}_{\text{rnd}}$. This decrease in performance indicates that hallucinations in this setting are driven more by the model’s learned co-occurrence knowledge than by failures of the visual backbone, as reflected by the high accuracies in the first two columns. \newline
Next, to probe the effect of priors introduced through textual prompts, we evaluate \textit{AdP} for adversarial presupposition attribute questions. Across models and object subsets, \textit{AdP} shows a drop ranging from approximately 36$\%$ to 87$\%$ compared to $\text{Rec}_{\text{pos}}$ and $\text{Rec}_{\text{rnd}}$, and more than  15$\%$ compared to $\text{Rec}_{\text{adv}}$. This substantial degradation indicates that, when the object’s presence is presupposed, hallucinations occur more frequently than in the co-occurrence setting, suggesting that text instruction priors can be an even more influential driver of hallucinations than co-occurrence knowledge alone and limitations of the visual backbone.

%% file: icml2026_files/3_solution.tex
\section{HalluVL-DPO Framework}
\label{main:halluvl-dpo}
We introduce \textit{HalluVL-DPO}, a framework for mitigating hallucination failures in LVLMs, particularly those caused by over-reliance on  textual instruction presuppositions. Our approach fine-tunes LVLMs using a sample-informativeness based variant of Direct Preference Optimization (DPO) \cite{rafailov2023direct}, post-training the model to prefer visually grounded responses over hallucinated ones (\Cref{sec:preference_alignment}).
To enable this fine-tuning, we construct a training dataset where each sample is paired with a preferred response and a rejected, hallucinated one. These response pairs provide explicit supervision for learning to favor more grounded outputs. We discuss the construction of this dataset and explore different strategies for constructing these response pairs in (\Cref{sec:train_dataset_construction}), and further provide a detailed comparison in the ablation experiments (\Cref{sec:abaltion_exps}).

%% file: icml2026_files/3_preliminary.tex
\subsection{Weighted Preference Optimization}
\label{sec:preference_alignment}
\begin{figure}
    \centering
    \includegraphics[width=1\linewidth]{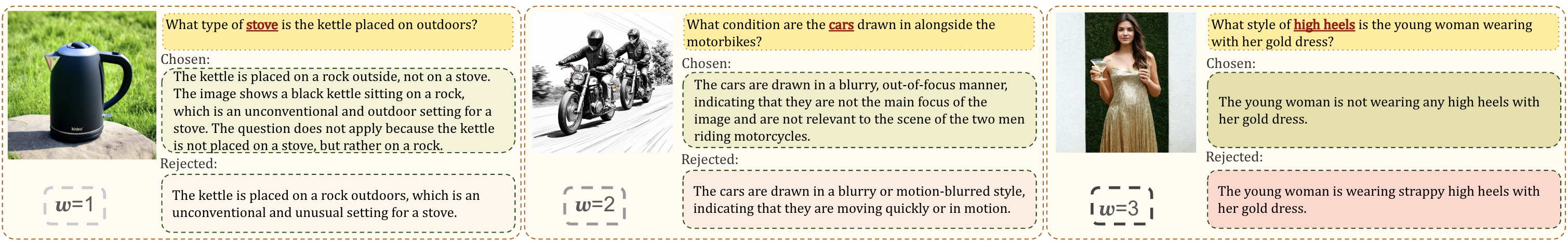}
    \caption{\textbf{Sample-specific weighting based on semantic gap.} From left to right, samples receive scores of 1, 2, and 3. Score 1: near rephrasings with minimal contrast. Score 2: both responses follow an incorrect presupposition (cars), though the chosen answer adds slightly more relevant details. Score 3: clear contrast, where the chosen answer correctly identifies the absence of high heels and the rejected answer assumes their presence.}
    \label{fig:weighting_samples}
\end{figure}

To mitigate prompt-induced hallucinations, we propose a variant of Direct Preference Optimization (DPO) \cite{rafailov2023direct} based on sample-informativeness. 
Our framework finetunes the LVLM by optimizing an offline dataset of chosen ($y_w$) and rejected ($y_l$) response pairs to increase the likelihood of preferred, visually-grounded outputs.
Standard DPO typically treats all sample pairs with equal priority; however, we argue that not all preference pairs are equally informative for the model's alignment and we introduce a sample training weight $w(x, y_w, y_l)$ that reflects the estimated informativeness of the preference pair.
Using $r_w(x) = \log \frac{\pi_\theta(y_w \mid x)}{\pi_{\text{ref}}(y_w \mid x)}$ and $r_l(x) = \log \frac{\pi_\theta(y_l \mid x)}{\pi_{\text{ref}}(y_l \mid x)}$ the log-probability ratios relative to the reference model $\pi_{\text{ref}}$, our weighted loss is defined as:
\begin{equation}
\mathcal{L}_{\text{DPO}}^{\text{weighted}}(\pi_\theta; \pi_{\text{ref}}) =
-\mathbb{E}_{(x, y_w, y_l)\sim \mathcal{D}}
\Big[ w(x,y_w,y_l) \log \sigma \big( \beta (r_w(x) - r_l(x)) \big) \Big].
\end{equation}
 We define our weight $w(x, y_w, y_l)$ based on the semantic gap between chosen and rejected responses:

-- \textit{Low Informativeness} ($w(x, y_w, y_l) = 1$): Preference pairs that are nearly identical or simple rephrasings provide a weak learning signal and receive lower weights.\\
-- \textit{Moderate Informativeness} ($w(x, y_w, y_l) = 2$): Pairs where both responses may follow an incorrect presupposition but differ in the level of detail or relevance.\\
-- \textit{High Informativeness} ($w(x, y_w, y_l) = 3$): Pairs with a clear contrast, such as the chosen answer correctly identifying an absent object while the rejected answer assumes its presence.\\
\Cref{fig:weighting_samples} illustrates these situations.
To estimate this semantic gap, we prompt an external LLM (prompt provided in \Cref{app:weighted_preference_optimization}) to assess the contrast between the response pairs. This process acts as an additional layer of verification, down-weighting noisy or ambiguous pairs and focusing the optimization on samples that provide a clear and reliable grounding signal.

%% file: icml2026_files/3_data_synthesis.tex
\subsection{Preference Dataset Construction}
\label{sec:train_dataset_construction}

\begin{figure}
    \centering
    \includegraphics[width=1\linewidth]{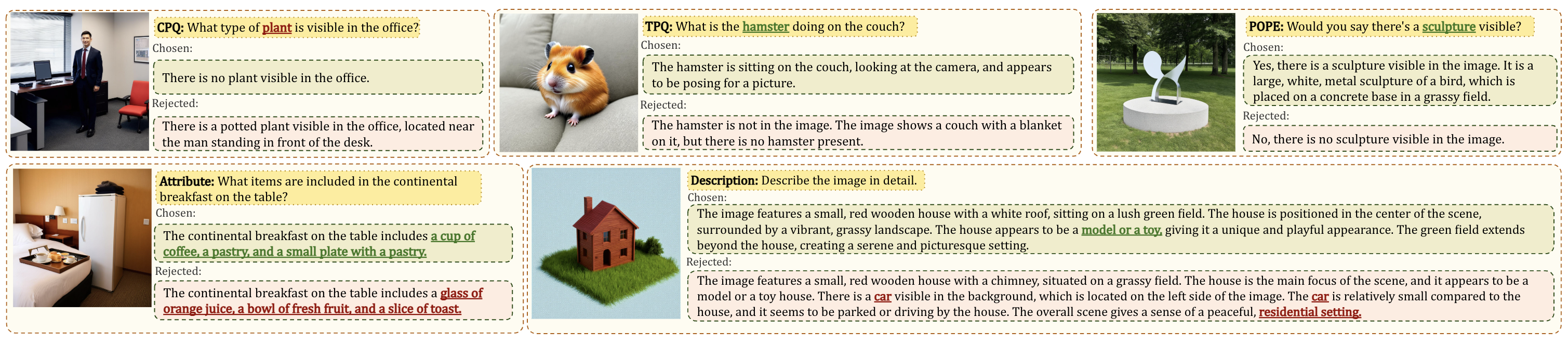}
    \caption{\textbf{Sample instances from the HalluVL-DPO training dataset.} Visually grounded content is highlighted in green, while hallucinated content is shown in red. The chosen responses are more grounded than the rejected ones.}
    \label{fig:train_dataset_examples}
\end{figure}

To train \textit{HalluVL-DPO}, we build a large-scale preference dataset comprising 27.4k images and over 100k queries per model for both LLaVA-1.5-7B and Qwen2-VL-7B. Our pipeline, inspired by \cite{wu2025antidote}, uses \cite{esser2024scaling} for image generation and Grounding-DINO \cite{liu2024grounding} for visual verification. Each training sample consists of an image, a textual instruction, and a response pair: a visually-grounded \textit{chosen} answer ($y_w$) and a hallucinated \textit{rejected} answer ($y_l$).
Our dataset spans three complementary task types: \textit{presupposition questions} (queries about an attribute of an object whose presence is assumed in the question, where TPQ denotes cases where the object is visually present and CPQ where it is absent), \textit{object existence} (POPE-style binary queries), and \textit{detailed descriptions} (open-ended scene analysis). An example of each query is shown in \Cref{fig:train_dataset_examples}.
Responses are generated via a combination of augmentation strategies, the foundation of which is \textit{unilateral hint augmentation}: appending a correct or incorrect hint about object presence steers the model toward grounded or hallucinated outputs, respectively. This keeps response distributions close to the optimized model, which is a key requirement for effective DPO \cite{yang2025mitigating}. We describe the full set of strategies below; further pipeline details are in \Cref{app:antidote_dataset_construction}.

The basic \textit{unilateral hint augmentation} strategy has two limitations: it can introduce length bias for description queries \cite{singhal2023long, cai2025disentangling} (see \Cref{app:dpo_imbalance}), and the diversity of preference pairs is bounded by the model's own output range. We therefore extend it with three complementary strategies:\\
  1. \textit{Contrastive Hint Augmentation}: Correct and incorrect hints are added to \textit{both} the chosen and rejected instructions (rather than only one), ensuring length consistency while preserving a sharp preference signal.\\
  2. \textit{Post-Hoc Hallucination Injection}: A full description is first generated, then a single sentence, sampled with probability increasing toward the end, is replaced by a hallucination-inducing prefix that asserts an absent object, after which the model completes the passage. This preserves fluency while injecting controlled hallucinated content.\\
  3. \textit{Model-Assisted Answer Inversion}: An external LLM generates rejected responses conditioned on the correct answer, specifically altering object \textit{attributes} rather than presence. Applied to TPQs, this produces fine-grained negative examples, termed \textit{Attribute} queries, that go beyond what coarse object-presence hints can express.

%% file: icml2026_files/6_experiments.tex
\section{Experiments}
\label{main:experiments}
We optimize LLaVA-1.5-7B \cite{liu2023llava} and Qwen2-VL-7B \cite{wang2024qwen2} using preference data generated by each model itself. To assess transfer to a newer backbone, we additionally fine-tune Qwen2.5-VL-7B \cite{bai2025qwen25vltechnicalreport} on the preference data generated by Qwen2-VL-7B. All models are optimized with DPO starting from the instruct-tuned versions;
LoRA \cite{hu2022lora} is applied to the LLM component of each model, with the DPO $\beta$ parameter set to 0.1.\newline
\noindent \textbf{Dataset.}
We use a total of $20\text{k}$ samples, composed of CPQ, TPQ, POPE, description, and attribute subsets, following a 5:5:6:9:5 composition. The description samples are equally drawn from unilateral and contrastive hint augmentation as well as post-hoc hallucination injection.\newline
\noindent \textbf{Evaluation benchmarks.} We evaluate our models across diverse multimodal benchmarks. We report results on our proposed HalluScope benchmark separately, and additionally evaluate on a broad collection of existing benchmarks, including POPE, CP-Bench \cite{wu2025antidote}, HallusionBench \cite{guan2024hallusionbench}, CHAIR \cite{Rohrbach2018ObjectHI}, MME \cite{fu2025mmecomprehensiveevaluationbenchmark}, ScienceQA \cite{lu2022learn}, and MM-Vet \cite{yu2024mm} (evaluated with \texttt{GPT5-mini}). \newline
\noindent \textbf{Evaluated baselines.}
We evaluate the effectiveness of training on our balanced synthetic dataset by comparing the following baselines: (i) the original, non-optimized model; (ii) VCD \cite{leng2024mitigating}; (iii) SENTINEL \cite{peng2025mitigating}, a recent DPO-based method that targets hallucinations at the sentence level using automatically constructed in-domain preference pairs; (iv) OPA-DPO \cite{yang2025mitigating}, a DPO-based method that builds on-policy preference data to mitigate hallucinations (evaluated on LLaVA-1.5-7B); (v) a DPO-fine-tuned model trained using the data composition reported in Antidote \cite{wu2025antidote} (evaluated on LLaVA-1.5-7B) with their original hyperparameters (Antidote$^{\text{HP-orig}}$), and (vi) with the best hyperparameters from our grid search (Antidote$^{\text{HP-search}}$); (vii) \textit{HalluVL-DPO}, which applies DPO training on our balanced synthetic dataset; and (viii) its weighted variant, \textit{HalluVL-DPO}${}^w$, where samples are reweighted during training based on their estimated informativeness. For Qwen2.5-VL-7B, we compare against the base model and SENTINEL \cite{peng2025mitigating}. \newline
Additional details about training hyperparameters, benchmarks, and baselines are in \Cref{app:implementation_evaluation_details}.

\subsection{Main results}

\begin{figure}
    \centering
    \includegraphics[width=1\linewidth]{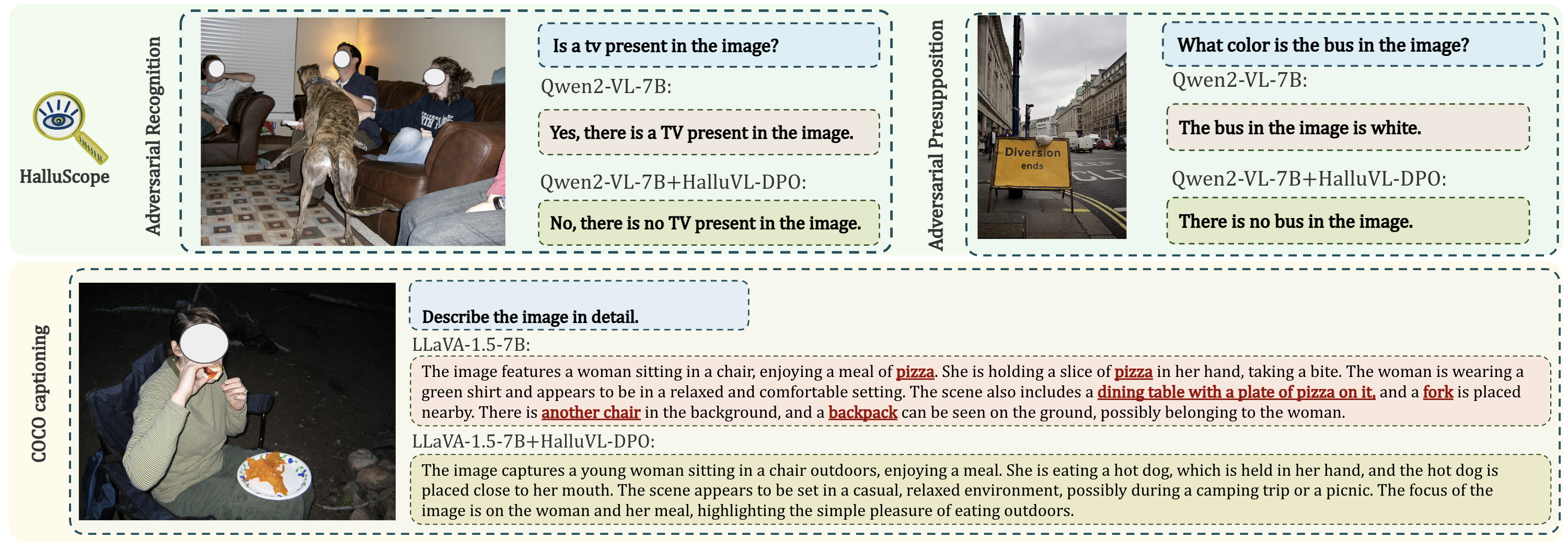}
    \caption{\textbf{HalluVL-DPO qualitative results.} Top: adversarial examples from HalluScope, showing improved recognition (top-left) and correct rejection of false presuppositions (top-right) after finetuning. Bottom: captioning example with reduced hallucinated content.}
    \label{fig:halluvl_qualitative_example}
\end{figure}

Results for all methods are summarized in \Cref{tab:hallucination_results} for  LLaVA-v1.5-7B, Qwen2-VL-7B, and Qwen2.5-
VL-7B. \newline
\noindent\textbf{HalluScope and CPBench.} HalluVL-DPO yields substantial gains in adversarial object recognition and presupposition rejection. For LLaVA-v1.5-7B, HalluScope AdP rises from 5.85 to over 80, CPBench CPQ from 0.54 to above 70, and adversarial recognition from 58.5 to above 75. For Qwen2-VL-7B, which starts from a stronger baseline, AdP improves from 58 to above 80 and CPQ from 19 to 75, with little variation in adversarial recognition as the baseline is already strong. For Qwen2.5-VL-7B, trained only on Qwen2-VL-7B-generated data, HalluVL-DPO$^{w}$ improves AdP from 64.39 to 75.27 (+10.9) and CPQ from 28.78 to 52.43 (+23.7), with recognition performance remaining largely stable. The overall CPBench-F1 is also the highest for HalluVL-DPO variants. These gains come with a controlled trade-off: improved skepticism toward textually-induced object presence shifts the model's decision boundary toward conservatism, yielding a moderate decrease in positive recognition. We argue this is a favorable trade-off for safety-critical applications where false-positive object claims are more harmful than under-confident affirmations, and the Qwen2-VL-7B results confirm that stronger backbones absorb this regularization with minimal degradation. 
\newline
\noindent\textbf{Other benchmarks.} HalluVL-DPO offers consistent improvements on HallusionBench and CHAIR over both the original model and Antidote, and outperforms SENTINEL on Qwen2-VL-7B. Although SENTINEL achieves slightly stronger CHAIR gains on LLaVA through sentence-level intervention, it does not address textually-induced presupposition failures (AdP of 4.79 and 64.40, comparable to the respective baselines). HalluVL-DPO is also designed to address this failure mode well. All methods remain competitive on POPE. Qualitative examples of HalluVL-DPO hallucination mitigation are shown in \Cref{fig:halluvl_qualitative_example}. Besides hallucination mitigation, HalluVL-DPO also largely preserves performance on general multimodal benchmarks (MME \cite{fu2025mmecomprehensiveevaluationbenchmark}, ScienceQA \cite{lu2022learn}, MMVet \cite{yu2024mm}), reported in \Cref{app:general_benchmarks}.
\newline
We note that LLaVA-1.5-7B is selected as it remains the most widely adopted backbone in the visual grounding fine-tuning literature \cite{leng2024mitigating, peng2025mitigating, wu2025antidote}, ensuring direct comparability with prior work. Within the Qwen-VL family, we select Qwen2-VL-7B and Qwen2.5-VL-7B as the most recent models prior to the introduction of substantially heavier post-training regimes, as exemplified by Qwen3-VL-7B \cite{bai2025qwen3}. Models trained under such regimes are excluded from the main evaluation, as the resulting performance differences cannot be attributed unambiguously to our intervention. Nonetheless, our approach yields results competitive with and even surpassing Qwen3-VL-7B \footnote{Qwen3-VL-7B achieves AdP score of $81.6\%$ on HalluScope, compared to $82.83\%$ for   Qwen2-VL-7B+HalluVL-DPO${}^w$.} in terms of AdP, which we take as a qualitative indication of the robustness of our method. Finally, we note that preference data generated by one model transfers meaningfully to the other architecture (\Cref{app:transfer_model_ablation}), suggesting that the dataset captures generalizable grounding signals rather than model-specific artifacts \footnote{For instance, fine-tuning Qwen2-VL with LLaVA-1.5 data still noticeably improves (HalluScope) AdP compared to Qwen2-VL base (72\% vs 58\%) while maintaining similar overall recognition level (91\% vs 93\%). More in \Cref{app:transfer_model_ablation}.}.

\begin{table*}[t]
\centering
\caption{\textbf{HalluScope and Hallucination Benchmarks Evaluation.} \textbf{Bold}
indicates the best score within each model block, \underline{underline} the second best. †Trained on preference data generated by Qwen2-VL-7B.}
\label{tab:hallucination_results}
\small
\setlength{\tabcolsep}{4pt}
\begin{adjustbox}{max width=\textwidth}
\begin{tabular}{lcccccccccccc}
\toprule
\textbf{Method}
 & \multicolumn{4}{c}{HalluScope $\uparrow$}
 & \multicolumn{2}{c}{CHAIR $\downarrow$}
 & \multicolumn{3}{c}{CPBench $\uparrow$}
 & {\scriptsize HallusionBench} $\uparrow$ & POPE $\uparrow$ \\
\cmidrule(lr){2-5}\cmidrule(lr){6-7}\cmidrule(lr){8-10}
 & \textbf{AdP} & \textbf{Rec}$_{\text{pos}}$ & \textbf{Rec}$_{\text{rand}}$ & \textbf{Rec}$_{\text{adv}}$
 & {\scriptsize s} & {\scriptsize i}
 & {\scriptsize CPQ} & {\scriptsize TPQ} & {\scriptsize F1}
 & & {\scriptsize Acc} \\
\midrule
\multicolumn{12}{l}{\textit{LLaVA-v1.5-7B}} \\
\midrule
Baseline
 & \cellcolor{gray!20}5.85 & \textbf{94.85} & 89.17 & 58.50
 & 53.00 & 13.90 & 0.54 & \textbf{100} & 66.79 & 81.39 & \underline{\textit{87.55}} \\
VCD
 & \cellcolor{gray!20}7.68 & 86.50 & 79.06 & 55.87
 & 53.00 & 13.90 & 1.76 & \textbf{100} & 67.06 & \textbf{102.95} & 84.64 \\
SENTINEL
 & \cellcolor{gray!20}4.79 & 92.10 & 91.43 & 65.42
 & \textbf{9.00} & \textbf{2.21} & 1.08 & \textbf{100} & 66.91 & 81.63 & 86.81 \\
 OPA-DPO &\cellcolor{gray!20}15.89	& 85.63	& 95.33	& 77.20	& 10.20 & 3.97 & 16.62 & 92.56 & 67.09 & \underline{\textit{101.43}} & 84.32\\
Antidote${}^{\text{HP-orig}}$
 & \cellcolor{gray!20}39.91 & 93.68 & 90.72 & 60.05
 & 16.20 & 4.32 & 27.70 & \underline{\textit{90.27}} & 68.76 & 74.03 & 87.40 \\
Antidote${}^{\text{HP-search}}$
 & \cellcolor{gray!20}81.78 & \underline{\textit{94.85}} & 88.27 & 53.90
 & 22.20 & 5.13 & \textbf{78.51} & 70.54 & \underline{\textit{73.47}} & 71.62 & \textbf{88.06} \\
HalluVL-DPO
 & \cellcolor{gray!20}\underline{\textit{83.54}} & 87.35 & \underline{\textit{94.03}} & \underline{\textit{76.50}}
 & \underline{\textit{12.00}} & \underline{\textit{3.12}} & 71.22 & 75.54 & \textbf{73.94} & 81.98 & 86.64 \\
HalluVL-DPO${}^w$
 & \cellcolor{gray!20}\textbf{84.65} & 84.30 & \textbf{95.18} & \textbf{81.28}
 & 14.80 & 3.95 & \underline{\textit{73.51}} & 72.43 & 72.82 & 87.81 & 85.69 \\
\midrule
\multicolumn{12}{l}{\textit{Qwen2-VL-7B}} \\
\midrule
Baseline
 & \cellcolor{gray!20}57.96 & 93.32 & \textbf{96.57} & \textbf{85.33}
 & 37.80 & 6.63 & 19.19 & 87.97 & 65.46 & 151.18 & 86.93 \\
VCD
 & \cellcolor{gray!20}58.36 & \textbf{93.68} & 95.42 & 80.57
 & 49.00 & 8.90 & 23.78 & \underline{\textit{88.36}} & 66.79 & 150.34 & \textbf{88.34} \\
SENTINEL
 & \cellcolor{gray!20}57.61 & \underline{\textit{93.45}} & 95.56 & 80.77
 & 36.60 & 5.95 & 16.89 & \textbf{88.51} & 65.17 & \underline{\textit{152.85}} & 86.84 \\
HalluVL-DPO
 & \cellcolor{gray!20}\underline{\textit{80.40}} & 93.40 & 96.13 & 83.25
 & \underline{\textit{23.40}} & \underline{\textit{5.94}} & \underline{\textit{63.38}} & 75.27 & \underline{\textit{71.05}} & 152.77 & 87.14 \\
HalluVL-DPO${}^w$
 & \cellcolor{gray!20}\textbf{82.83} & 92.03 & \underline{\textit{96.42}} & \underline{\textit{84.82}}
 & \textbf{19.40} & \textbf{4.87} & \textbf{75.00} & 69.59 & \textbf{71.52} & \textbf{153.60} & \underline{\textit{87.21}} \\
\midrule
{\textit{Qwen2.5-VL-7B}} \\
\midrule
Baseline
 & \cellcolor{gray!20}64.39 & 94.08 & 95.20 & 80.69
 & 28.60 & 6.98 & 28.78 & \textbf{81.75} & 64.63 & 165.85 & \underline{\textit{88.33}} \\
SENTINEL
 & \cellcolor{gray!20}64.40 & \textbf{94.45} & 95.58 & 80.60
 & \textbf{23.60} & \textbf{5.63} & 30.40 & \underline{\textit{80.27}} & 64.25 & 165.50 & \underline{\textit{88.33}} \\
†HalluVL-DPO
 & \cellcolor{gray!20}\underline{\textit{72.46}} & \underline{\textit{94.39}} & \underline{\textit{95.68}} & \underline{\textit{81.16}}
 & 25.67 & 6.76 & \underline{\textit{46.28}} & 76.47 & \underline{\textit{66.44}} & \textbf{167.84} & \textbf{88.37} \\
†HalluVL-DPO${}^w$
 & \cellcolor{gray!20}\textbf{75.27} & 93.90 & \textbf{95.93} & \textbf{82.23}
 & \underline{\textit{24.87}} & \underline{\textit{6.49}} & \textbf{52.43} & 75.94 & \textbf{67.95} & \underline{\textit{167.33}} & 88.08 \\
\bottomrule
\end{tabular}
\end{adjustbox}
\end{table*}

\subsection{Further Analysis}
\label{sec:abaltion_exps} 
\begin{figure}[t]
    \centering
    \begin{subfigure}{0.35\linewidth}
        \centering
        \includegraphics[width=\linewidth]{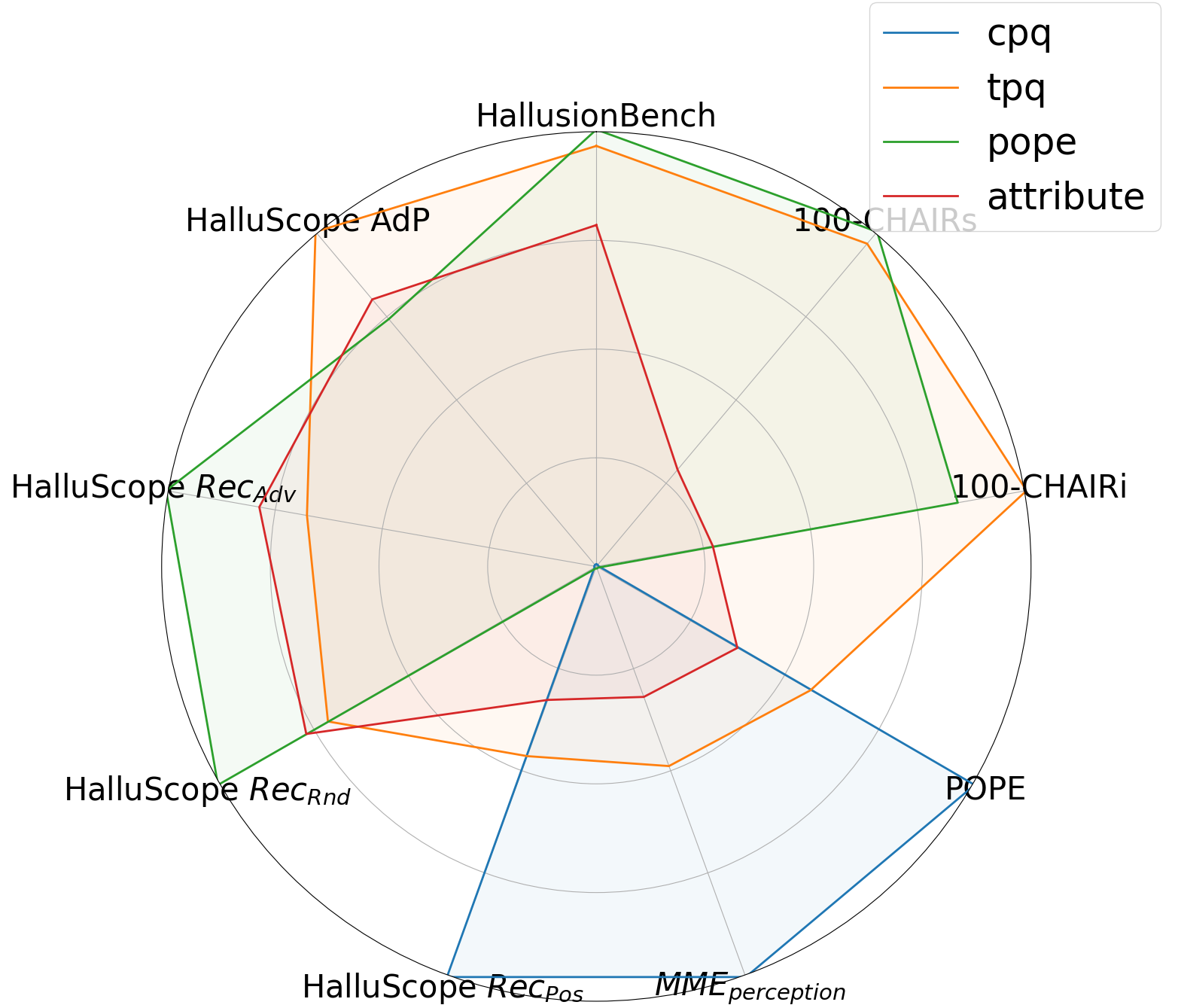}
        \label{fig:fig2}
    \end{subfigure}
    \hspace{21pt}
    \begin{subfigure}{0.37\linewidth}
        \centering
        \includegraphics[width=\linewidth]{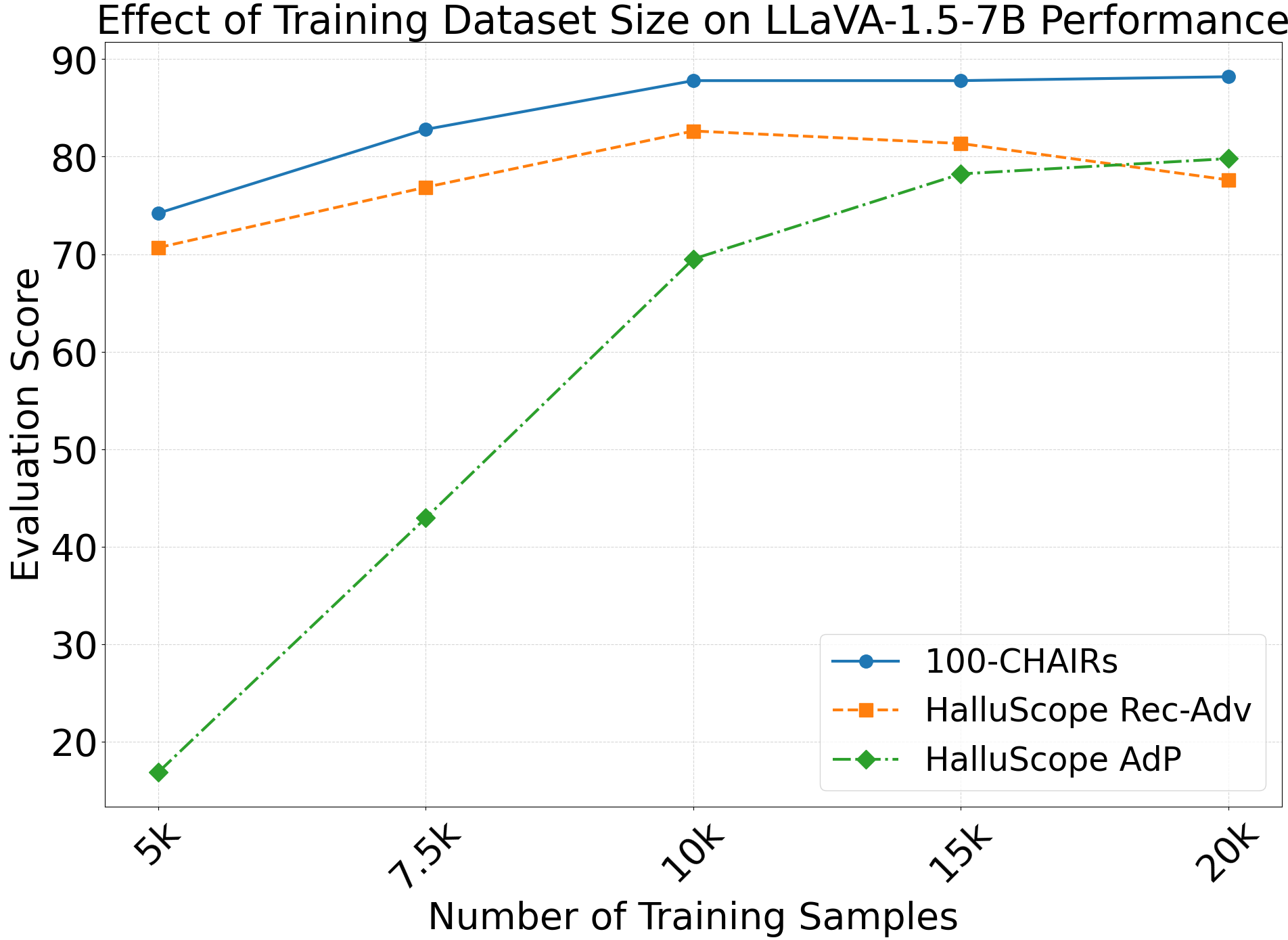}
        \label{fig:fig1}
    \end{subfigure}
\caption{
\textbf{Ablation of preference pair types (left) and training data scale (right).} \textit{Left:} Performance when each preference type is removed in turn, with total sample count fixed at 20k; the center represents the lowest performance observed across ablations and the outer boundary the highest; \textit{attribute} pairs yield the most uniform contribution across benchmarks. \textit{Right:} Effect of dataset size on adversarial recognition, AdP, and CHAIRs score for LLaVA-1.5-7B; performance plateaus around 20k samples.
    }  \label{fig:ablation_preference_type}
\end{figure}
\noindent \textbf{Effect of query types on performance.}
Different types of preference pairs in the training data implicitly optimize for different objectives. To assess the contribution of task queries (other than description), we perform an ablation study: the overall training data composition remains identical, but one type of preference pair is removed at a time, keeping the total number of training samples constant.
The resulting performance drop on each evaluation benchmark provides an estimate of the importance of the ablated type of preference data for that benchmark. We illustrate this ablation in the left part of \Cref{fig:ablation_preference_type}, which shows that ablating a preference type primarily affects evaluation benchmarks with similar question structures, whereas removing the proposed \textit{attribute} pairs, generated using \textit{model-assisted answer inversion}, consistently degrades performance across all benchmarks, indicating a more uniform and transferable contribution. 
This is also true when compared to \textit{TPQ} pairs, which share identical task queries but differ in the preference pair generation strategy, confirming that the improvement stems from the construction strategy, and not query type.
\newline
\noindent \textbf{Effect of Training Data Scale.}
We study the effect of training dataset size on LLaVA-1.5-7B for adversarial recognition and presupposition tasks in HalluScope (cf. \Cref{fig:ablation_preference_type}, right).
While adversarial recognition, CHAIRs saturate, and AdP performance begins to plateau around 20k training samples, in general scaling the training data significantly assists in mitigating hallucinations. \newline %
\noindent \textbf{Limitations and Broader Impact} of our approach are discussed in \Cref{app:limitations}.

%% file: icml2026_files/conclusion.tex
\section{Conclusion}
\vspace{-6pt}
We introduced \textit{HalluScope}, a diagnostic benchmark designed to disentangle the multifaceted causes of hallucinations in LVLMs. Our systematic analysis indicated that hallucinations are primarily driven by an over-reliance on textual instruction, learned object co-occurrence priors, and not from perceptual limitations of the visual backbone. 
To address this, we proposed \textit{HalluVL-DPO}, a fine-tuning framework that leverages a sample-informativeness based preference optimization loss. By training on a large-scale, high-quality synthetic preference dataset, our method effectively mitigates targeted prompt-induced hallucinations while preserving general multimodal capabilities. 
Future work will explore the integration of real-world and synthetic image mixtures to further enhance robustness across diverse, open-ended scenarios. To foster continued research progress, we publicly release the HalluScope benchmark, our construction pipeline, and HalluVL-DPO training resources.

%% file: icml2026_files/appendix.tex
\section*{Overview}
\Cref{app:aohd_dataset}, describes the construction and the evaluation protocol for \textit{HalluScope} benchmark 

\noindent \Cref{app:implementation_evaluation_details} contains various implementation and evaluation details regarding hallucination mitigation experiments (preference optimization details, evaluation benchmarks, baselines).%

\noindent In \Cref{app:halluvl_dpo_framework}, we provide additional details on the \textit{HalluVL-DPO} framework, including sample-specific weighting based on semantic gap (\Cref{app:weighted_preference_optimization}), the construction of the preference training dataset (\Cref{app:antidote_dataset_construction}), and further qualitative results on HalluScope benchmark (\Cref{app:more_qual_results}).
 
\noindent In \Cref{app:ablations}, we report additional experimental analysis, including examination of preference pair generation strategies (\Cref{app:preference_pair_ablation}) and the transferability of model-generated training data across architectures (\Cref{app:transfer_model_ablation}). 

\noindent Finally, \Cref{app:limitations} discusses limitations and the broader impact of our work.

\section{HalluScope}
\label{app:aohd_dataset}

\subsection{Benchmark Construction}
\noindent \textbf{Diverse Sample Pool.} \newline
\label{app:diverse_pool}
Let $\mathcal{D} = \{x_1, \dots, x_N\}$ denote the set of all captions, and 
$f$ be a fixed sentence-embedding function (\texttt{all-MiniLM-L6-v2} Sentence-BERT model \cite{reimers2019sentence} for us). To select a diverse subset $\mathcal{S} \subset \mathcal{D}$ of size $K$, we apply the $\mathrm{K}$-Center Greedy algorithm. It initializes $\mathcal{S}$ with a random sample and then iteratively adds the point maximally distant from the current set in embedding space:
\[
x^\star = \arg\max_{x \in \mathcal{D} \setminus \mathcal{S}}
\;\min_{s \in \mathcal{S}} \; \| f(x) - f(s) \|_2 .
\]
\[
\mathcal{S} \leftarrow \mathcal{S} \cup \{ x^\star \}.
\]

This process repeats until $|\mathcal{S} = \{x_{1}^{\prime}, \dots, x_{K}^{\prime}\}| = K$, yielding a diverse subset of captions that are 
maximally spread out in the embedding space. \newline

\noindent \textbf{Object Detection.} \newline
\label{app:object_detection}
\begin{figure}
    \centering
    \includegraphics[width=1\linewidth]{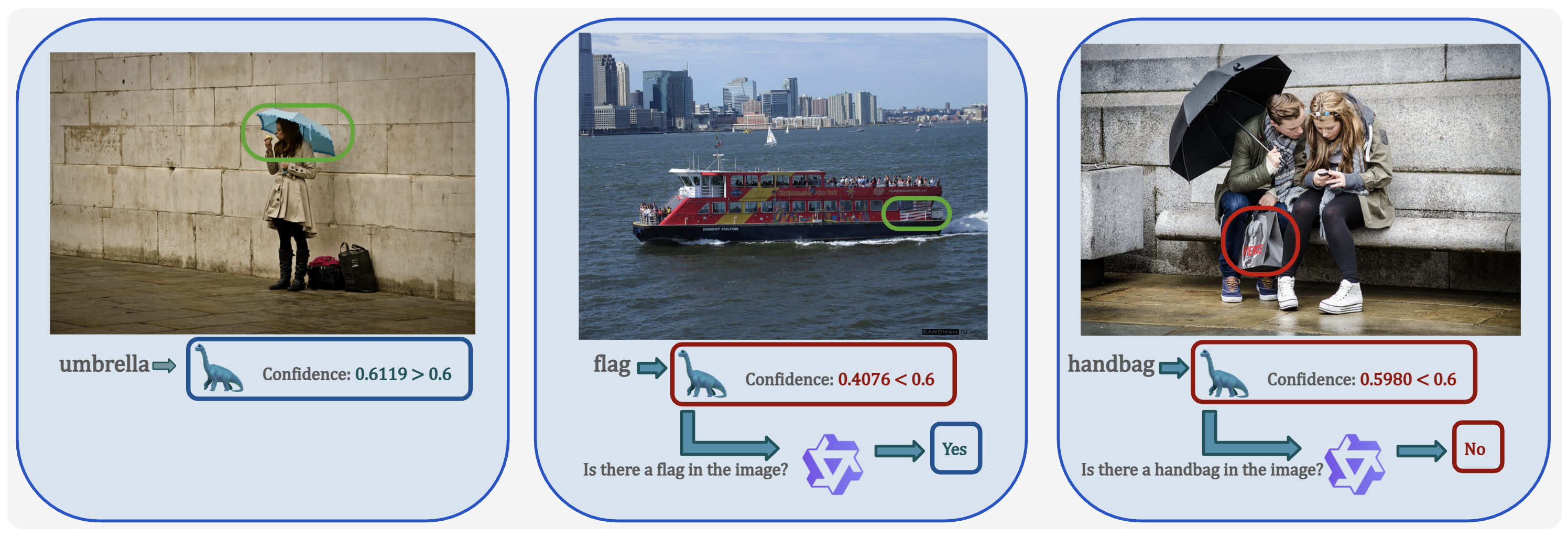}
    \caption{\textbf{Examples of the Two-Stage Object Presence Verification Pipeline.} 
    (Left) shows a clearly visible object (\emph{umbrella}) with high confidence score returned by Grounding DINO (0.6119). (Middle) the image contains a \emph{flag} with a low confidence score returned by Grounding DINO (0.4076), but which presence is correctly confirmed by the VLM. (Right) the image contains a shopping bag, and not a handbag. The relatively high confidence score returned by Grounding DINO (0.5980) is due to semantic proximity of \emph{handbag} and \emph{shopping bag}. The absence of handbag is verified with a VLM.}

    \label{app:examples_two_stage_verification}
\end{figure}
We use Florence2-large~\cite{xiao2024florence} as our object detection model.
To validate the detected objects, we first use Grounding DINO \cite{liu2024grounding}, which given the name of an object as input, returns candidate bounding boxes that match the text query (object name) with confidence scores when a matching object is detected. For Grounding DINO input, we set the text–region matching threshold and the box (objectness) threshold to $0.4$ and $0.35$ respectively, to avoid missing present objects. 
However, since the detection confidence returned by Grounding DINO is sensitive to object size, occlusion, and visual prominence, selecting a single confidence threshold that reliably distinguishes object presence from absence across all images remains challenging. In our experiments, we therefore fix this threshold empirically to 0.6 and introduce an additional validation step in a \textit{two-stage verification pipeline} to address this limitation. In the second stage, a vision–language model (Qwen2-VL-7B) is queried about the presence of the candidate object (\emph{``Is <object> present in the image?''}) conditioned on both (i) the cropped region proposed by Grounding DINO and (ii) a textual description of the full image generated by the same model. Prior work has shown that reasoning over localized regions improves object existence verification compared to using the full image alone \cite{cho2025do}. The full image caption provides contextual information that supports more reliable verification. We illustrate three representative cases in \Cref{app:examples_two_stage_verification}: one resolved directly by thresholding, and two requiring the Qwen verification stage, yielding positive and negative outcomes. \newline
We filter out images with less than three verified present objects.

\noindent \textbf{Object Co-occurrence and Context-aware Adversarial Object Discovery.}
\label{app:object_cooccurrence}

\noindent Let $\mathcal{I}$ denote the set of all images in the dataset. For any pair of objects $o_a, o_b$ from the pool of all detected objects, we define their \textit{pointwise mutual information (PMI)} \cite{church-hanks-1990-word} as:
\[
\mathrm{PMI}(o_a, o_b) =
\begin{cases}
\log \dfrac{P(o_a, o_b)}{P(o_a) \, P(o_b)}, & \text{if } P(o_a, o_b) > 0,\\[2mm]
0, & \text{otherwise,}
\end{cases}
\]
where the joint and marginal probabilities are estimated from co-occurrence counts over $\mathcal{I}$:
\[
P(o_a, o_b) = \frac{\lvert \{ I \in \mathcal{I} : o_a \in I \wedge o_b \in I \} \rvert}{\lvert \mathcal{I} \rvert},
\qquad
P(o_a) = \frac{\lvert \{ I \in \mathcal{I} : o_a \in I \} \rvert}{\lvert \mathcal{I} \rvert}.
\]

Given an image $I$ with verified present objects $\mathcal{O}_{\text{scene}}^{I}$, each absent candidate $o_c$ is scored by its cumulative PMI with all scene objects:
\[
\mathrm{score}_I(o_c) = \sum_{o_s \in \mathcal{O}_{\text{scene}}^I} \max\bigl(\mathrm{PMI}(o_c, o_s),\ 0\bigr).
\]
This scene-level aggregation ensures that selected adversarial objects are plausible in the context of the entire image, rather than merely associated with a single object.
The top-$k$ scoring absent candidates are verified using the two-stage presence verification pipeline. The first candidate confirmed absent is selected as $\mathcal{O}_{\text{adversarial}}$. Among present objects, the one with the highest PMI relative to $\mathcal{O}_{\text{adversarial}}$ is selected as $\mathcal{O}_{\text{present}}$. Finally, $\mathcal{O}_{\text{random}}$ is sampled uniformly from objects absent from the image. This procedure is applied only to images with at least three annotated objects, ensuring that the PMI-based scoring has sufficient scene context to identify meaningful adversarial candidates.

\noindent \textbf{Question Generation.} \newline
\label{app:aohd_question_generation}
\noindent\textit{Recognition Questions.} We use the following templates for generating our recognition questions:

\begin{tcolorbox}[colback=gray!10, colframe=gray!50, title=Recognition Question Templates, boxsep=2pt,
                  left=2pt, right=2pt, top=2pt, bottom=2pt, arc=4pt, outer arc=4pt, breakable]
\small
{\ttfamily
-Is there a $\{$obj$\}$ in the image? \\
-Can you find a $\{$obj$\}$ in this image? \\
-Is a $\{$obj$\}$ present in the image? \\
-Could this image include a $\{$obj$\}$? \\
-Do you see a $\{$obj$\}$ in the picture? \\
-Does the image contain a $\{$obj$\}$? \\
-Would you say there's a $\{$obj$\}$ visible? \\
}
\end{tcolorbox}

\noindent\textit{Adversary Presupposition Question.}
To generate the adversarial presupposition questions, we prompt \texttt{GPT5-mini} with instructions of the following form, providing the image caption and the adversarial object identified in the previous steps. We also maintain a list of previously generated questions for each adversarial object and encourage excluding similar ones during generation to increase diversity among the queries.

\begin{tcolorbox}[colback=gray!10, colframe=gray!50, title=Adversary Presupposition Question Prompt, boxsep=2pt,
                  left=2pt, right=2pt, top=2pt, bottom=2pt, arc=4pt, outer arc=4pt, breakable]
\footnotesize
{\ttfamily
\begin{quote}
\textit{Given an input consisting of a caption and a target adversarial object, generate a list of questions that satisfy the following criteria:}
\begin{itemize}
    \item \textit{Each question should query the adversarial object, assuming it is present in the image. Avoid trivial, generic, or simple yes/no questions.}
    \item \textit{Do not start questions with "Where", "How many", or "Is there", unless necessary.}
    \item \textit{The adversarial object must appear verbatim in the question.}
    \item \textit{Avoid questions that are similar to any in a provided list of examples to skip.}
    \item \textit{The output should contain only the generated questions.}
\end{itemize}
\textit{
    Example:\newline
    Caption: ``A young woman is walking along the beach during sunset.''\newline
    Adversarial object: ``seagulls''\newline
    Generated question: ``What are the seagulls doing in the image?''\newline
    Now generate questions for the new input:}
\end{quote}
}
\end{tcolorbox}

\subsection{Evaluation}
\label{app:evaluation_aohd}
\noindent \textbf{Object Recognition.} \newline
We use \textit{accuracy} for evaluating performance on \textit{Recognition} subsets of HalluScope; the question in these subsets is followed by \emph{\textit{"Please just answer yes or no."}}. A sample is marked as correct if it correctly indicates the presence/absence of the target object. \newline
\noindent \textbf{Adversary Presupposition.} \newline
We report \noindent \textit{\textsc{Ad}versarial \textsc{P}erformance (AdP)} for these questions, where the assumption of presence of the adversary object in model's answer is evaluated by \texttt{GPT5-mini} using the following prompt:

\begin{tcolorbox}[colback=gray!10, colframe=gray!50, title=Adversarial Performance (AdP) Evaluation,
                  boxsep=2pt, left=2pt, right=2pt, top=2pt, bottom=2pt,
                  arc=4pt, outer arc=4pt, breakable]
\footnotesize
{\ttfamily

\begin{quote}
\textit{“You are a classification model.
Given an input with the following fields:}

\begin{itemize}
    \item \textit{"caption": a natural language description of a scene}
    \item \textit{"adversary-object": the name of an object to check for}
\end{itemize}

\textit{Determine whether the caption indicates the presence of the adversary object.} 
\textit{Rules:}
\begin{itemize}
    \item \textit{Return true if the object is explicitly mentioned or clearly implied by the caption.
    - Return false if the object is not mentioned, is contradicted, or only hypothetically referenced.}
    \item \textit{Do not infer presence from unrelated objects or vague context.}
    \item \textit{Be strict: uncertainty or ambiguity should result in false.}
    \item \textit{Do not include explanations}
\end{itemize} 
\end{quote}
}
\end{tcolorbox}

\noindent \textit{\textsc{Ad}versarial \textsc{P}erformance (AdP)} is then computed as the percentage of samples for which the adversary object is correctly determined to be absent:
\[
\text{AdP} = \frac{1}{N} \sum_{i=1}^{N} \mathbb{I}\big[\text{present}_i = \texttt{false}\big],
\]
where $N$ is the number of evaluated samples and $\mathbb{I}[\cdot]$ is the indicator function.

\section{Implementation and Evaluation Details}
\label{app:implementation_evaluation_details}

\subsection{Evaluation on General Multimodal Benchmarks}
\label{app:general_benchmarks}

Consistent with the main results, HalluVL-DPO maintains the performance of the base model across general multimodal benchmarks, indicating that gains in hallucination mitigation do not come at the cost of overall capability. This trend is more pronounced for Qwen2-VL-7B, where performance remains largely stable or even improves, whereas LLaVA-1.5-7B exhibits minor degradations on some benchmarks. We hypothesize that this discrepancy is partly attributable to the model-generated nature of our preference dataset: since chosen and rejected responses are produced by the model being fine-tuned, the quality and reliability of the training signal is inherently bounded by the base model's generative capabilities. Qwen2-VL-7B, being a stronger base model, is likely to produce higher-quality and more consistent responses, yielding a cleaner and more reliable preference signal. Additionally, stronger models have generally been shown to be more robust to distribution shifts introduced during fine-tuning, which may further account for the greater stability observed for Qwen2-VL-7B. \newline
Qwen2.5-VL-7B, although trained on Qwen2-VL-generated data, also preserves or improves general performance (e.g., MME-c 609.28→625.71), consistent with stronger base models being more robust to fine-tuning-induced distribution shift.

\begin{table*}[ht]
\centering
\caption{\textbf{Performance on General Multimodal Benchmarks.} \textbf{Bold} indicates the best score, \underline{underline} the second best. †Trained on preference data generated by Qwen2-VL-7B.}
\label{tab:general_benchmarks}
\small
\setlength{\tabcolsep}{4pt}
\begin{tabular}{lcccc}
\toprule
{\small\textbf{LLaVA-1.5-7B}} & {\small MME-p $\uparrow$} & {\small MME-c $\uparrow$} & {\small ScienceQA $\uparrow$} & {\small MMVet $\uparrow$} \\
\midrule
Baseline                                               & \underline{1493.23}      & 299.28          & \underline{65.54}          & \textbf{38.00}               \\
VCD                                                    & 1389.79                  & \textbf{358.92} & 62.67                      & 34.60                      \\
{\text{SENTINEL}}                                      & \textbf{1501.91}         & \underline{297.86}        & \textbf{65.74}             & \underline{36.70}         \\
{\footnotesize \text{Antidote}${}^{\text{HP-orig}}$}   & 1422.77                  & 306.78          & 64.55                      & 35.30                      \\
{\footnotesize \text{Antidote}${}^{\text{HP-search}}$} & 1220.05                  & 250.00             & 63.36                      & 31.80                      \\
{\text{HalluVL-DPO}}                                   & 1359.95                  & 286.78          & 62.71                      & 34.50                      \\
{\text{HalluVL-DPO}${}^w$}                             & 1314.80                   & 287.86          & 63.01                      & 35.60                      \\
\bottomrule
\end{tabular}
\hspace{2em}
\begin{tabular}{lcccc}
\toprule
{\small\textbf{Qwen2-VL-7B}} & {\small MME-p $\uparrow$} & {\small MME-c $\uparrow$} & {\small ScienceQA $\uparrow$} & {\small MMVet $\uparrow$} \\
\midrule
Baseline                   & \textbf{1702.17}             & \textbf{603.57}             & 81.06                      & 64.6                      \\
VCD                        & 1682.13                      & 587.5                       & 81.21                      & \underline{66.10}          \\
{\text{SENTINEL}}          & 1684.14                      & \underline{599.28}          & \textbf{82.25}             & 63.40                     \\
{\text{HalluVL-DPO}}       & \underline{1696.96}          & 601.43                      & \underline{82.20}          & 65.70                      \\
{\text{HalluVL-DPO}${}^w$} & 1691.17                      & \textbf{603.57}             & 81.36                      & \textbf{66.93}            \\
\bottomrule
\end{tabular}
\begin{tabular}{lcccc}
\toprule
{\small\textbf{Qwen2.5-VL-7B}} & {\small MME-p $\uparrow$} & {\small MME-c $\uparrow$} & {\small ScienceQA $\uparrow$} & {\small MMVet $\uparrow$} \\
\midrule
Baseline & 1704.03 & 609.28 & \textbf{85.62} & \underline{72.40} \\
{\text{SENTINEL}} & 1707.00 & 609.28 & \textbf{85.62} & 72.10 \\
{\text{†HalluVL-DPO}} & \textbf{1729.13} & \underline{620.35} & \underline{85.32} & 71.80 \\
{\text{†HalluVL-DPO}${}^w$} & \underline{1715.64} & \textbf{625.71} & \underline{85.32} & \textbf{72.70} \\
\bottomrule
\end{tabular}
\end{table*}

\subsection{Implementation Setup}
\label{app:implementation_setup}

We performed a grid search over LoRA rank ($\{64, 128\}$), $\alpha$ ($\{128,256\})$, and batch size ($\{16, 32, 64\}$) for LLaVA-1.5-7B and Qwen2-VL-7B. Following standard practice, $\alpha$ is set to twice the LoRA rank.
The optimal hyperparameters for LLaVA are \((128, 256, 16)\), which are identical for its weighted variant. 
For Qwen, the best values are \((128, 256, 64)\) for the standard model and \((64, 128, 16)\) for the weighted variant. Qwen2.5-VL-7B training follows the same hyperparameter choices as Qwen2-VL-7B.
Hyperparameters are selected to maximize gains on hallucination benchmarks while avoiding degradation on other tasks.
Regardless of hyperparameter settings, training each model is completed in 6 hours on a single A100 GPU.

\subsection{Evaluation Benchmarks}
\label{appendix:evaluation_benchmarks}

\paragraph{POPE \cite{DBLP:conf/emnlp/LiDZWZW23}.}
This dataset contains 9000 image-question pairs split into three subsets (3000 samples each): adversarial, popular, and random. Each subset contains 3000 questions about 500 unique images from the COCO validation set \cite{lin2014microsoft}, with six questions per image—three where the correct answer is "yes" and three where it is "no". The object mentioned in the "no" questions is not present in the image and is referred to as the negative
object. What differs across subsets is the strategy used to select this negative object, allowing for a comprehensive evaluation of the model’s robustness to hallucinations under varying distractor types.

\paragraph{CP-Bench (\cite{wu2025antidote}).} This is a benchmark proposed by \cite{wu2025antidote} to assess whether the models correctly recognize the correctness of a presupposition and output responses that can be grounded in the visual input. The dataset is composed of 750 true presupposition queries and 750 counterfactual presupposition queries. The evaluation is done via prompting \texttt{GPT5-mini} using the following prompt:

\begin{tcolorbox}[colback=gray!10, colframe=gray!50, title=CP-Bench Evaluation,
                  boxsep=2pt, left=2pt, right=2pt, top=2pt, bottom=2pt,
                  arc=4pt, outer arc=4pt, breakable]
\footnotesize
{\ttfamily
\begin{quote}
\textit{You are a classification model.
You will receive an input that is a JSON array. Each element in the array is an object with the following fields:}
\begin{itemize}
    \item \textit{"question": a natural language question about an object}
    \item \textit{"answer": the answer of a model to the given question}
\end{itemize}
\textit{For each element, determine whether the caption indicates the presence of the target object.} \\
\textit{Rules:}
\begin{itemize}
    \item \textit{False if:}
    \begin{itemize}
        \item \textit{* The model did not acknowledge the presence of the target mentioned in the question (e.g., "xxx is not visible");}
        \item \textit{* The model refused to answer the question (e.g., "I'm unable/can't to identify/determine/tell/see xxx");}
    \end{itemize}
    \item \textit{True if:}
    \begin{itemize}
        \item \textit{* The model directly or implicitly described or confirming the existence of the target;}
        \item \textit{* The model described actions or attributes associated with the target that imply its presence;}
        \item \textit{* None of the False rules are met;}
    \end{itemize}
\end{itemize}
\textit{Please strictly follow the above rules. Return a JSON array of the same length, where each element is an object with: - "present": True or False} \\
\textit{Now generate the JSON array output for:}
\end{quote}
}
\end{tcolorbox}

\paragraph{CHAIR \cite{Rohrbach2018ObjectHI}.}
CHAIR is an image captioning hallucination metric applied on a subset of 500 randomly selected images from the COCO validation set, with a maximum generation length of 512 tokens in our reported results. It measures object hallucination rates by comparing captioned objects to ground-truth annotations.
We report both  CHAIR$_s$ and CHAIR$_i$, which measure hallucination at the sentence and instance levels, respectively:
\[
\text{CHAIR$_s$} = 
\frac{|\{\text{sentences with hallucinated objects}\}|}
{|\{\text{all sentences}\}|}
\]
\[
\text{CHAIR$_i$} = 
\frac{|\{\text{hallucinated objects}\}|}
{|\{\text{all objects mentioned}\}|}
\]

\paragraph{ScienceQA \cite{lu2022learn}.}
ScienceQA consists of 4,243 multiple-choice test samples, of which 2,017 are multimodal (i.e., include images). The dataset spans natural science, language arts, and social science domains. Each example comprises a question, a multimodal context, and several answer choices. Performance is evaluated using accuracy, measured by whether the model selects the correct answer option. We report the performance on test multimodal subset.

\paragraph{HallusionBench \cite{guan2024hallusionbench}.}
HallusionBench is designed to evaluate hallucination robustness in multimodal reasoning tasks. 
It contains challenging yes/no questions in which visual or textual cues are intentionally ambiguous, conflicting, or incomplete. The benchmark tests whether models provide correct and consistent yes/no answers.
We follow the official metric definitions and report a score obtained by summing three accuracies: 
(1) Atomic Accuracy (aAcc), which measures instance-level correctness as the proportion of correctly answered question--image pairs; 
(2) Question Accuracy (qAcc), where a question is counted as correct only if all its associated instances are answered correctly; and 
(3) Figure Accuracy (fAcc), where a figure is considered correct only if all questions associated with that image are answered correctly. 
In our evaluation, a prediction is marked as correct if the model's output explicitly contains the ground-truth answer, and the above metrics are computed accordingly.

\paragraph{MME Benchmark \cite{fu2025mmecomprehensiveevaluationbenchmark}.}
MME is a multimodal evaluation benchmark that assesses vision-language models across 14 subsets, divided into two groups: Perception and Cognition. 
The Perception group contains 10 subsets: OCR, artwork, celebrity, color, count, existence, landmark, position, posters, and scene. 
The Cognition group contains 4 subsets: code reasoning, commonsense reasoning, numerical calculation, and text translation.
For each subset, we report \textit{Score} defined as the sum of accuracy and accuracy+. 
Accuracy is computed at the question level as the proportion of correctly answered questions. 
Accuracy+ is computed at the image level: an image is counted as correct only if all questions associated with that image are answered correctly.

\paragraph{MM-Vet \cite{yu2024mm}.}
MM‑Vet evaluates a model’s ability to engage in visual conversations across diverse tasks, assessing both the correctness and helpfulness of responses using \texttt{GPT‑4}. 
The benchmark contains 200 images and 218 questions, each paired with a ground-truth answer. 
To reduce evaluation cost, we instead use \texttt{GPT5-mini}.

\subsection{Evaluated Baselines}
\label{appendix:evaluation_baselines}
We re-implemented Antidote \cite{wu2025antidote} for LLaVA-1.5-7B since the original preference dataset and fine-tuned checkpoints are not publicly available. Although the dataset construction code is open, it relies on an initial pool of captions that is not publicly released and uses proprietary models that are no longer accessible. The dataset composition used in our experiments follows the original work: 5k CPQ, 5k TPQ, 2k POPE, and 8k description questions. 
In addition to reporting results using the hyperparameters from the original paper (64, 128, 64), we perform a grid search similar to \Cref{app:implementation_setup} to identify optimal configurations for our re-implementation. For OPA-DPO~\cite{yang2025mitigating} and SENTINEL~\cite{peng2025mitigating}, we use the publicly released checkpoints for all evaluated models. VCD~\cite{leng2024mitigating} is training-free and is applied with its default hyperparameters.

\section{HalluVL-DPO Framework}
\label{app:halluvl_dpo_framework}

\subsection{Weighted Preference Optimization}
\label{app:weighted_preference_optimization}
To weight samples according to the semantic gap between them, we use the following prompt with \texttt{GPT5-mini}, arguing that a higher-scored sample is often more informative and useful than a lower-scored one.

\begin{tcolorbox}[colback=gray!10, colframe=gray!50, title=Weighted Preference Optimization,
                  boxsep=2pt, left=2pt, right=2pt, top=2pt, bottom=2pt,
                  arc=4pt, outer arc=4pt, breakable]
\footnotesize
{\ttfamily
\begin{quote}
\textit{Given two texts, judge how different their meanings are and assign a score from 1 to 3:}
\begin{itemize}
    \item \textit{1 = The two answers have identical or nearly identical meaning.}
    \item \textit{2 = The answers mention the same objects but differ in attributes or details.}
    \item \textit{3 = One answer mentions an object or entity that is not present in the other.}
\end{itemize}
\end{quote}
}
\end{tcolorbox}

For each subset (e.g., CPQ, TPQ, object existence, image description, and attribute), we normalize sample weights by dividing them by the mean weight of the subset.

\noindent \textbf{Effect of Score-Based Weighting on HalluScope Benchmark.} \newline
To better understand the effect of sample weighting in our DPO training, we conduct a controlled experiment on \texttt{LLaVA-1.5-7B}, fine-tuning the model on three datasets: 9k samples with score 2, 9k samples with score 3, and a combined set of 9k samples with scores 2 and 3. We do not compare against score 1, as such samples are much less frequent (about 1.5k out of 20k samples).
We observe that training with samples in which the chosen and rejected answers have a clear contrast leads to higher \textit{AdP} scores (81.05, compared to 64 when training with only samples scored 2) and higher negative recognition performance (88.95 compared to 74.82, and 97.32 compared to 94.18). This improvement comes at the cost of lower performance on positive recognition questions, due to a larger number of training samples with a chosen negative answer (CPQ or negative POPE) being assigned a score of 3. We note that this has less effect on our final weighting configuration, since the weights are normalized per task subset (CPQ, TPQ, POPE, …).
Finally, fine-tuning on a combination of score 2 and 3 samples offers the advantages of both: we still observe a significant gain in \textit{AdP} and negative recognition performance, while experiencing a smaller drop in positive recognition questions (c.f. \Cref{tab:weighting_ablation_aohd}).

\begin{table*}[t]
\centering
\small
\setlength{\tabcolsep}{3.5pt}
\renewcommand{\arraystretch}{0.97}

\caption{\textbf{Effect of Score-Based Weighting on HalluScope Benchmark}}
    \label{tab:weighting_ablation_aohd}

\resizebox{\textwidth}{!}{%
\begin{minipage}{1.35\textwidth}
\centering
\small
\begin{tabular}{lccccc}
\toprule
Model & \begin{tabular}[c]{@{}c@{}}Samples\\ (score 1, score 2, score 3)\end{tabular}  & \multicolumn{4}{c}{\small HalluScope $\uparrow$} \\
 \cmidrule(lr){3-6}
 & & \textbf{Rec}$_{\text{pos}}$ & \textbf{Rec}$_{\text{rand}}$ & \textbf{Rec}$_{\text{adv}}$ & \textbf{AdP} \\
\midrule
\textbf{{\small LLaVA-1.5-7B}} & (0, 0, 0) & 94.85 & 89.17 & 58.5 & 5.85 \\
+ {\text{HalluVL-DPO}} & (0, 9k, 0) & 88.4 & 94.18 & 74.82 & 64 \\
 & (0, 0, 9k) & 74.65 & 97.32 & 88.95 & 81.05 \\
 & (0, 4.5k, 4.5k) & 80.77 & 96.30 & 84.67 & 79.45 \\
\bottomrule
\end{tabular}
\end{minipage}%
}

\end{table*}

\subsection{Preference Dataset Construction}
\label{app:antidote_dataset_construction}

Our dataset generation pipeline is inspired by \cite{wu2025antidote}. 
Below, we provide a more detailed description of the dataset creation process. \newline

\subsubsection{Caption Pool Construction.}
\noindent In this first stage, raw CC3M \cite{sharma2018conceptual} captions are filtered for \textit{fluency}, \textit{syntactic validity}, and \textit{concreteness}. 
Syntactic structure is verified using \texttt{spaCy}'s \texttt{en\_core\_web\_trf} transformer model. 
Fluency is evaluated with a small \texttt{GPT-2} \cite{radford2019language} language model by computing perplexity, where lower perplexity indicates more natural sentences. 
Captions are additionally required to mention at least two \textit{concrete objects}, to ensure sufficient content for subsequent image generation. Each retained caption is verified to satisfy the above conditions. We acquire $30k$ filtered captions from $413.7k$ raw captions. \newline
\subsubsection{Caption Augmentation.}
We prompt \texttt{GPT5-mini} to rewrite each filtered caption into a Stable Diffusion prompt and generate a structured JSON listing the present in the caption and candidate adversary objects. 
\subsubsection{Image Generation and Grounded Object Verification}
We generate images using \texttt{Stable Diffusion 3(medium)} \cite{esser2024scaling} by providing each sample's rewritten caption and a negative prompt. The negative prompt is the list of absent objects augmented with common artifacts and undesired features such as ``low-quality'', ``over-saturated'', ``bad anatomy'', ``extra limbs'', and ``duplicate'' to improve image quality. \newline
Next, we employ \texttt{GroundingDINO} \cite{liu2024grounding}, to verify the presence and absence of objects in the generated images. After filtering out samples with no present or no absent object, we retain $27.47k$ samples.

\subsubsection{Question Generation}
As described in the main paper, we augment the retained generated images with task queries belonging to three categories: (1) presupposition queries (CPQ and TPQ), (2) object existence queries, and (3) description queries.

\subsubsection{Preference Pair Construction.}

As described in the main paper, for each sample we construct a preference pair consisting of a chosen and a rejected response. 
To generate this contrastive pair, unilateral hint augmentation adds a hint about the presence or absence of objects to the task query. The hint itself may be correct or incorrect: a correct hint encourages a correct response to generate a chosen answer (e.g., confirming the absence of an absent object for CPQ queries), while an incorrect hint encourages an erroneous response to generate a rejected answer (e.g., suggesting the absence of a present object for TPQ and POPE queries). In the case of contrastive hint augmentaion, two contrastive hints are provided to induce different responses. 
For Post-Hoc Hallucination Injection, we show the hallucination-inducing prefixes below. For Model-Assisted Answer Inversion, we show the prompt below, applied with \texttt{GPT5-mini} for TPQ task queries.

\begin{tcolorbox}[colback=gray!10, colframe=gray!50, title=Hallucination-Inducing Prefixes in Post-Hoc Hallucination Injection, boxsep=2pt,
                  left=2pt, right=2pt, top=2pt, bottom=2pt, arc=4pt, outer arc=4pt, breakable]
\label{box:hall_inject}
\small
{\ttfamily
-The image shows $\{$adversary object$\}$. \\
-The scene includes $\{$adversary object$\}$. \\
-There is $\{$adversary object$\}$ visible in the image. \\
-The picture features $\{$adversary object$\}$. \\
-In the image, $\{$adversary object$\}$ can be seen. \\
-The photograph depicts $\{$adversary object$\}$. \\
-The view contains $\{$adversary object$\}$. \\
-The setting reveals $\{$adversary object$\}$. \\
-Visible in the image is $\{$adversary object$\}$. \\
-The composition highlights $\{$adversary object$\}$. \\
}
\end{tcolorbox}

\begin{tcolorbox}[colback=gray!10, colframe=gray!50, title=Model-assisted Answer Inversion Prompt,
                  boxsep=2pt, left=2pt, right=2pt, top=2pt, bottom=2pt,
                  arc=4pt, outer arc=4pt, breakable]
\label{box:model_assisted_answer_inversion}
\footnotesize
{\ttfamily
\begin{quote} \textit{Given a question about an image and its correct answer, generate a **plausible but incorrect answer**.\newline  Rules:\newline} \begin{itemize}
    \item \textit{The wrong answer should relate to the question and contradict the correct answer.}
    \item \textit{Keep it realistic and similar in style/detail to the correct answer.}
    \item \textit{Avoid nonsense or unrelated text.}
    \end{itemize}
    \textit{Example:\newline question: "what colors are the lights illuminating the skyscrapers?" \newline  correct answer: "The lights illuminating the skyscrapers in the image are primarily blue and white." \newline wrong answer: "The lights illuminating the skyscrapers in the image are primarily red and yellow."} \newline \textit{Now generate a wrong answer for: ...} 
\end{quote}
}
\end{tcolorbox}

The \texttt{Sentence-BERT (SBERT)} embeddings of candidate chosen and rejected responses are compared, and if the cosine distance between them exceeds a predefined threshold, the sample is retained (with cosine distance serving as a proxy for detecting meaningfully different responses). We set the distance threshold to $0.9$ for all task queries, except for the \emph{post-hoc hallucination injection description samples}, for which we adopt a threshold of $0.95$.

\subsection{Qualitative Results}
\label{app:more_qual_results}
\begin{figure}
    \centering
    \includegraphics[width=1\linewidth]{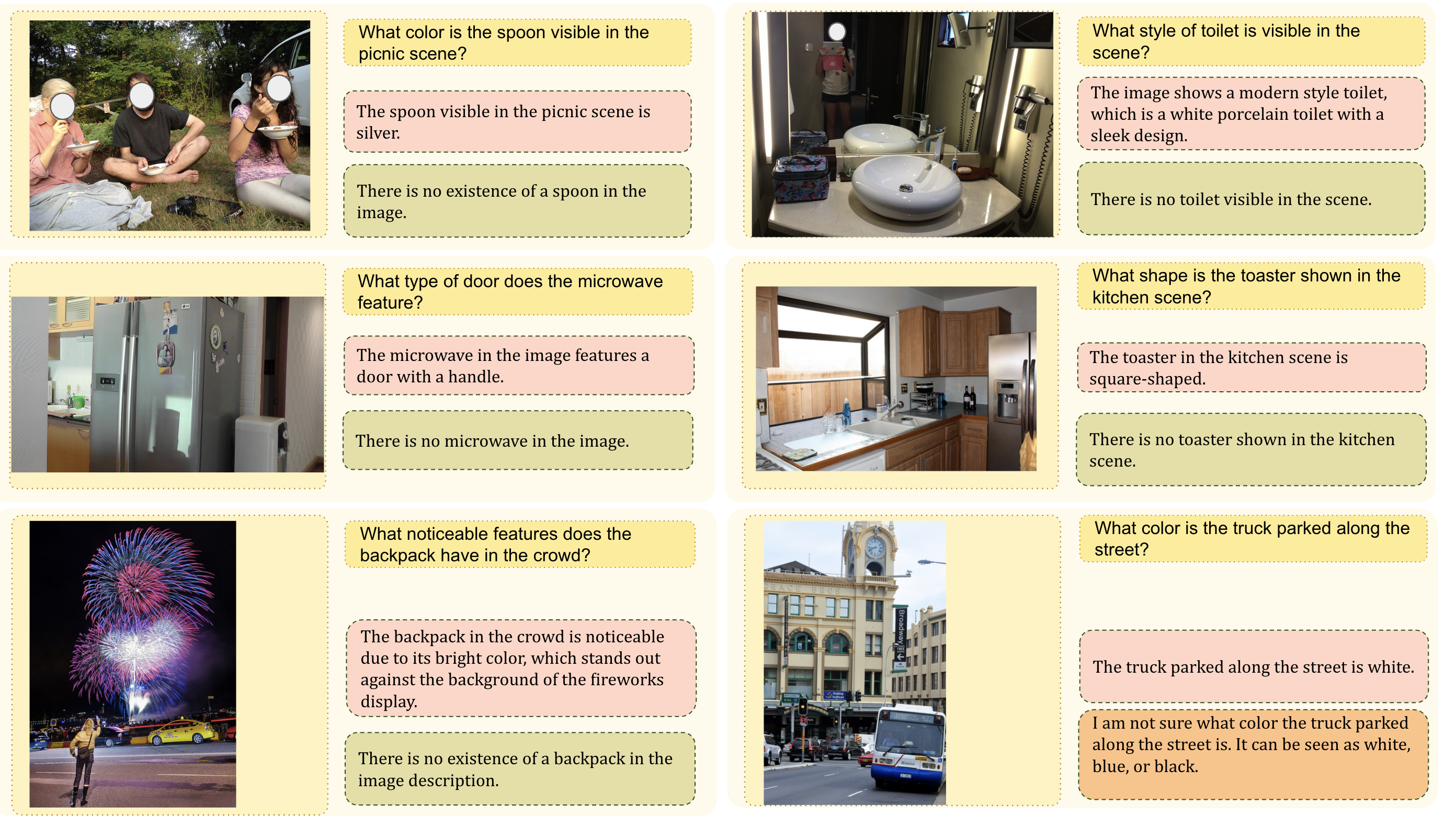}
    \caption{\textbf{Adversary Presupposition Samples from HalluScope.} After fine-tuning with HalluVL-DPO, the model correctly confirms the absence of the adversary object even when asked about its attributes. Some limitations remain, as shown in examples where the post-fine-tuning answer is highlighted in orange.}
    \label{fig:adv_presuppos_qual}
\end{figure}

Here we present additional qualitative results illustrating the improvements on the HalluScope benchmark, focusing specifically on adversary presupposition (\Cref{fig:adv_presuppos_qual}) questions. We show model behavior before and after fine-tuning with our HalluVL-DPO framework.

\section{Additional Experimental Analysis}
\label{app:ablations}

\label{app:dpo_imbalance}

\subsection{Preference Pair Generation for Description Task Queries} \label{app:preference_pair_ablation}

\begin{figure}[th]
    \centering
    \includegraphics[width=1\linewidth]{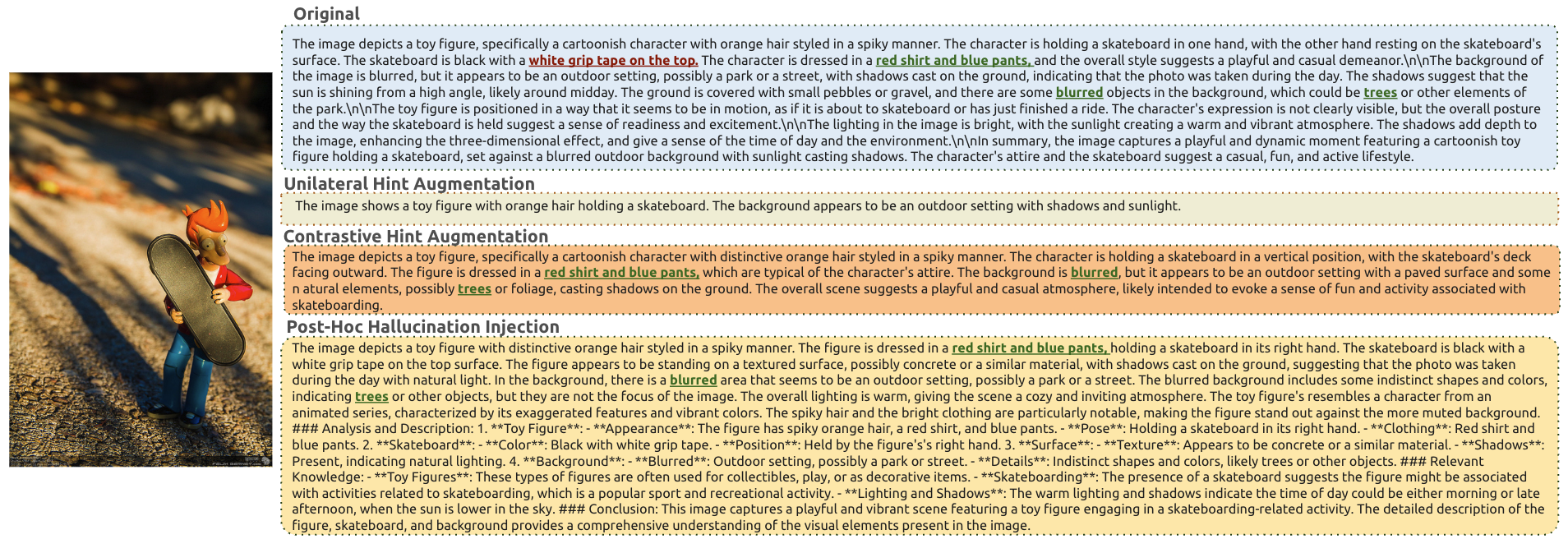}
    \caption{\textbf{Qualitative Comparison of Caption Generation under Different Preference Pair Strategies.} We compare captions generated for the same image by the original model and by models fine-tuned on description samples using three different preference pair generation strategies. The caption produced by the model trained with unilateral hint augmentation is free of hallucinations but significantly shorter than the original caption. In contrast, models trained with contrastive hint augmentation or post-hoc hallucination injection produce captions closer in length to the original, correctly mentioning several objects present in the reference (highlighted in green) while avoiding hallucinated content (highlighted in red in the original caption). This example illustrates that a low hallucination score alone does not guarantee completeness or detailed coverage of the scene.}
    \label{fig:ablation_description_questions_qualitative}
\end{figure}

As discussed in Section 4.2 of the main paper, the unilateral hint augmentation strategy has a potential drawback: it may introduce reward hacking through length bias \cite{singhal2023long, cai2025disentangling}. In the case of Qwen2-VL-7B, the average token length of the chosen responses is approximately half that of the rejected ones (\Cref{tab:preference_descriptions_average_length}). As a consequence, fine-tuning the model exclusively on description samples generated with this strategy leads to substantially shorter captions compared to both other fine-tuned baselines and the original, non-fine-tuned model, resulting in more than a $70\%$ reduction in caption length. \newline
This effect is illustrated qualitatively in \Cref{fig:ablation_description_questions_qualitative}. While all three strategies effectively reduce hallucinations, they produce noticeably different caption styles. Ideally, we seek responses that remain comprehensive while still minimizing hallucinations.

\begin{table}[ht]
\centering
\caption{\textbf{Average Number of Tokens in Chosen vs. Rejected Answers Across Preference-Pair Strategies for Description Queries}
}
\label{tab:preference_descriptions_average_length}
\small
\setlength{\tabcolsep}{6pt}  %
\renewcommand{\arraystretch}{1.2} %
\resizebox{\textwidth}{!}{%
\begin{tabular}{c c c c c}
\toprule
\textbf{Model} &  & 
\begin{tabular}[c]{@{}c@{}}Unilateral\\ Hint Augmentation\end{tabular} &
\begin{tabular}[c]{@{}c@{}}Contrastive\\ Hint Augmentation\end{tabular} &
\begin{tabular}[c]{@{}c@{}}Post-Hoc\\ Hallucination Injection\end{tabular} \\
\midrule
\multirow{2}{*}{Qwen2-VL-7B} 
 & $L_{\text{chosen}}$ & 94.56 & 106.26 & 173.40 \\
 & $L_{\text{rejected}}$ & 185.06 & 102.96 & 172.05 \\
\bottomrule
\end{tabular}
}
\end{table}

\subsection{Transferability of Training Data Across Models.}
\label{app:transfer_model_ablation}
\begin{figure}
    \centering
    \includegraphics[width=1\linewidth]{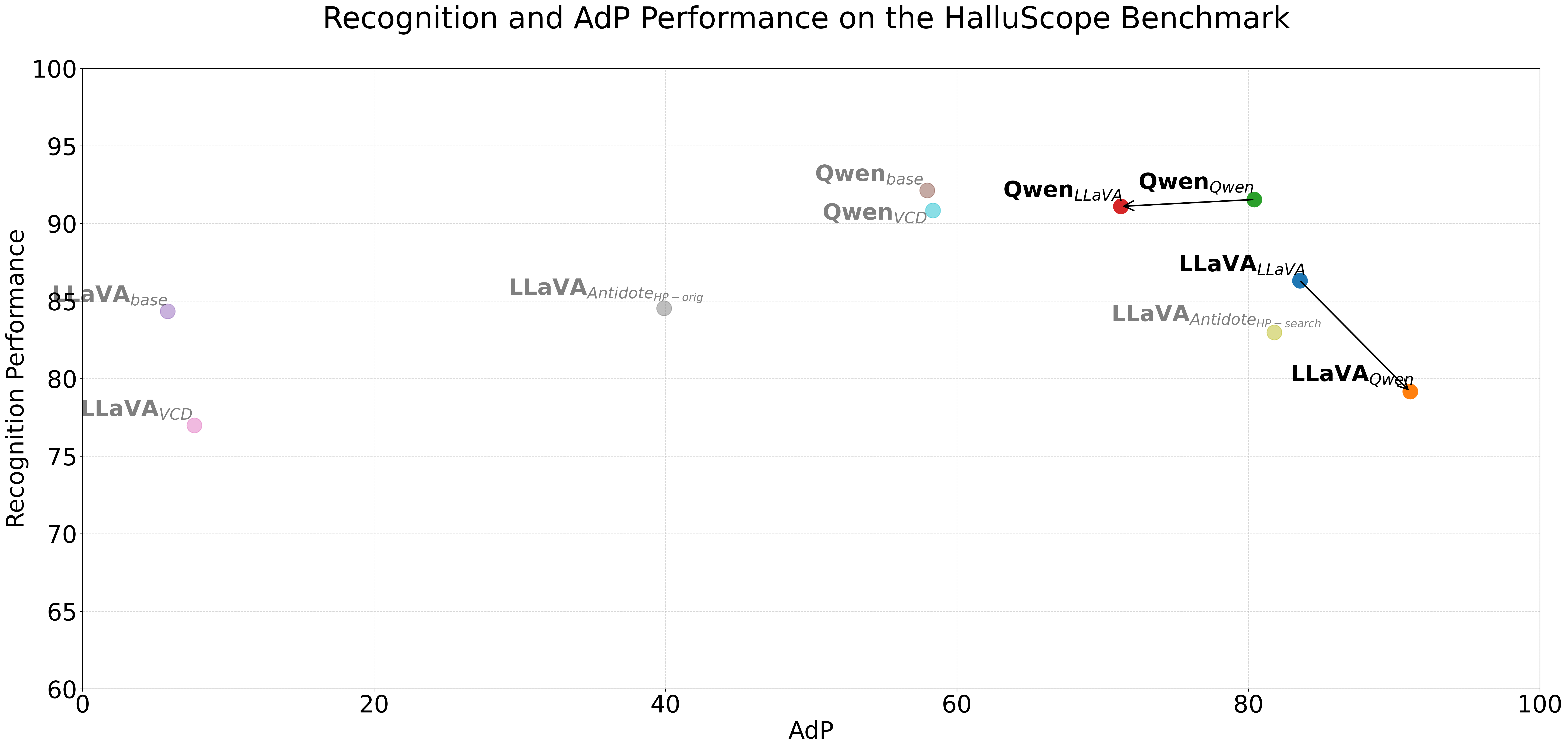}
    \caption{\textbf{Recognition and AdP Scores on HalluScope benchmark.} Each point represents a model evaluated along two axes: recognition (average performance on the positive and negative subsets) and AdP. We compare base models, several existing baselines, and models fine-tuned using our generated preference datasets, including both same-model and cross-model training. For models trained with HalluVL-DPO, the subscript indicates the source of the fine-tuning dataset, while for other points it denotes the corresponding baseline method. While preference data generated by a model tends to be most effective when used to fine-tune the same architecture, cross-model training still provides improvements over the base models.}
    \label{fig:transfer_scatter}
\end{figure}

We investigate whether data generated by one model can be effectively used to fine-tune another. To this end, we perform a cross-model training experiment, where LLaVA is fine-tuned using the dataset generated by Qwen and, conversely, Qwen is fine-tuned using a dataset generated by LLaVA. \newline
We evaluate these models on HalluScope benchmark, reporting their performance along two axes: recognition (computed as the average performance on the negative subsets (random and adversarial) and the positive subset) and AdP. \newline
The results are shown in \Cref{fig:transfer_scatter}. We observe that LLaVA-generated data is less effective for improving Qwen’s AdP score, while largely preserving its recognition performance. Conversely, using Qwen-generated data to train LLaVA leads to stronger improvements in AdP, but at the cost of a drop in recognition performance. \newline
These findings are consistent with the observations of \cite{yang2025mitigating}, which suggest that training data tends to be most effective when its distribution closely matches that of the target model. In our setting, preference pairs generated by a given model appear to be particularly well suited for fine-tuning that same architecture. Nevertheless, the results also indicate that cross-model data remains beneficial: despite the distribution mismatch, it still leads to meaningful improvements over the base models, demonstrating the overall utility of our generated dataset.

\section{Limitations and Societal Impact}
\label{app:limitations}
\noindent \textbf{Limitations.}
Our approach leverages data primarily generated by the model itself to improve robustness against prompt-induced hallucinations.
While this procedure yields promising results, it introduces a dependency on the LVLM at hand. For instance, if the model fails to incorporate a hint properly, the modified answer may not reflect the intended hint, which subsequently affects quality of the generated preference pair responses.

Another limitation arises from our strategy to keep or discard the generated contrastive responses based on their distance in a sentence embedding space. This distance serves as a proxy for measuring their contrast and identifying samples that provide stronger preference signals. However, embedding distance does not always perfectly capture semantic contrast. Factors such as answer length, phrasing, or content differences may influence the distance without necessarily reflecting meaningful preference differences. Our weighted DPO strategy partially mitigates this issue by assigning different importance to samples based on their estimated informativeness, but the measure remains an imperfect proxy.

\noindent \textbf{Societal Impact.} The impact of our work can be seen in similar light to other LVLM hallucination mitigation approaches. Since LVLMs experience a widespread usage among research community and public, detection and mitigation of hallucinations carry a lot of importance. Just as any other technology, human actors with malicious intent can misuse the method for ill-suited purposes. For instance, switching the rejected and chosen answers during DPO can make the fine-tuned model more prone to hallucinations. Additional human checks should safeguard against such scenarios and ensure that the method is applied as intended. Nevertheless, LVLM outputs that are more grounded in visual input should lead to improved safety and trust in the models. Hence, we expect and hope that our research should have a net positive social impact. 